%% file: main_camera_ready.tex
\documentclass{article}

\usepackage[eandd, final]{neurips_2026}

\usepackage[utf8]{inputenc}
\usepackage[T1]{fontenc}
\usepackage{hyperref}
\usepackage{url}
\usepackage{booktabs}
\usepackage{amsfonts}
\usepackage{nicefrac}
\usepackage{microtype}
\usepackage{xcolor}

\usepackage{amsmath,amssymb}
\usepackage{graphicx}
\usepackage[export]{adjustbox}
\usepackage{tabularx}
\usepackage{multirow}
\usepackage{makecell}
\usepackage{subfigure}
\usepackage{enumitem}
\usepackage{xspace}
\usepackage[most]{tcolorbox}
\usepackage{float}
\usepackage{placeins}
\usepackage{cleveref}

\usepackage{todonotes}
\usepackage{xcolor}
\usepackage{etoc}
\usepackage{wrapfig} 

\newcommand{\method}{TorchUMM\xspace}
\addtocontents{toc}{\protect\setcounter{tocdepth}{-1}}

\begin{document}

\title{TorchUMM: A Unified Multimodal Model Codebase for Evaluation, Analysis, and Post-training}

\author{%
  Yinyi Luo\textsuperscript{1}\thanks{Contact author: \texttt{yinyil@andrew.cmu.edu}.} \quad
  Wenwen Wang\textsuperscript{1} \quad
  Hayes Bai\textsuperscript{2} \quad
  Hongyu Zhu\textsuperscript{2} \quad
  Hao Chen\textsuperscript{1} \\
  \textbf{Pan He\textsuperscript{3}} \quad
  \textbf{Marios Savvides\textsuperscript{1}} \quad
  \textbf{Sharon Li\textsuperscript{4}} \quad
  \textbf{Jindong Wang\textsuperscript{2}}\thanks{Corresponding author: \texttt{jdw@wm.edu}.} \\[0.5em]
  \textsuperscript{1}Carnegie Mellon University \quad
  \textsuperscript{2}William \& Mary \\
  \textsuperscript{3}Auburn University \quad
  \textsuperscript{4}University of Wisconsin-Madison
}

\maketitle

\begin{abstract}

Recent advances in unified multimodal models (UMMs) have led to a proliferation of architectures capable of understanding, generating, and editing across visual and textual modalities.
However, developing a unified framework for UMMs remains challenging due to the diversity of model architectures and the heterogeneity of training paradigms and implementation details.
In this paper, we present \textbf{\method}, the \textit{first} unified codebase for comprehensive evaluation, analysis, and post-training across diverse UMM backbones, tasks, and datasets.
TorchUMM supports a broad spectrum of models covering a wide range of scales and design paradigms.
Our benchmark encompasses three core task dimensions: multimodal understanding, generation, and editing, and integrates both established and novel datasets to evaluate perception, reasoning, compositionality, and instruction-following abilities.
By providing a unified interface and standardized evaluation protocols, TorchUMM enables fair and reproducible comparisons across heterogeneous models and fosters deeper insights into their strengths and limitations, facilitating the development of more capable unified multimodal systems.
Code is available at: \url{https://github.com/AIFrontierLab/TorchUMM}.

\end{abstract}

\section{Introduction}

Recent advances in unified multimodal models (UMMs) have significantly expanded the scope of artificial intelligence systems, enabling a single model to process and reason over multiple modalities such as vision, language, and beyond~\citep{zhao2025unified, chen2025janus, xie2025show}.
These models have demonstrated strong capabilities across diverse tasks, including multimodal understanding~\citep{shao2025large, zhang2024vision}, generation~\citep{wang2024genartist, song2024moma, koh2023generating}, and complex reasoning~\citep{fei2024multimodal, wang2024genartist, wang2024exploring}.
Existing research primarily focuses on improving model architectures \citep{xie2024show, deng2025emerging, xie2025show, chen2025janus, wu2024janus, ma2024janusflow}, analyzing how performance scales with data and parameters \citep{li2025uni}, and exploring post-training techniques \citep{wang2026quantifying,qin2025uni, xie2025reconstruction, su2025unigame, han2026unicorn}.

Despite substantial progress, there is a lack of a unified evaluation solution for UMMs since their performances are often benchmark-dependent and inconsistent across tasks \citep{srivastava2023beyond, ouyang2022training, munjal2026instruction, feuer2024style}.
This suggests that many reported improvements are localized rather than indicative of a holistic enhancement in model capability.
On the other hand, existing evaluation protocols typically rely on a limited set of benchmarks, often focusing on a single task or capability \citep{mcintosh2025inadequacies, chen2024mmstar, zhang2024vision}.
Such practices can lead to overly optimistic conclusions.
Furthermore, differences in evaluation pipelines, data preprocessing, and model interfaces make it challenging to conduct consistent cross-model and cross-method comparisons.
These limitations highlight the need for a comprehensive, systematic, and extensible framework that goes beyond isolated benchmarks and single-task assessment.

\begin{figure*}[t!]
  \centering
  \includegraphics[width=.95\linewidth]{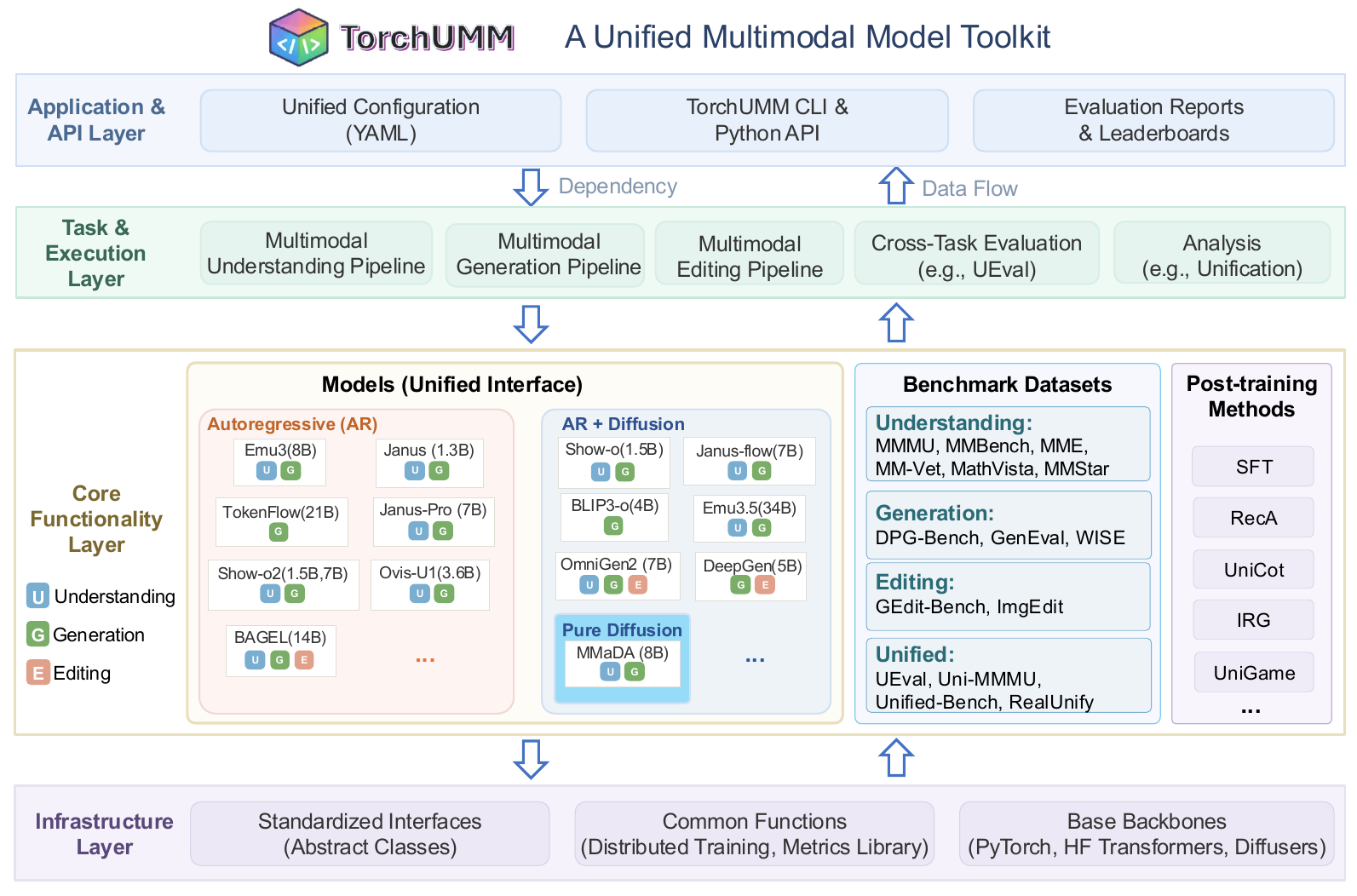}
  \vspace{-.1in}
  \caption{Overview of \method.}
  \label{fig:pipeline}
  \vspace{-.2in}
\end{figure*}

In this work, we present \textbf{\method} (\Cref{fig:pipeline}), the \textit{first} unified toolkit for systematic evaluation, analysis, and post-training of UMMs.
\method standardizes evaluation pipelines across models and tasks, enabling controlled and reproducible comparisons under consistent settings.
Moreover, it supports flexible integration of diverse post-training methods that supports multi-dimensional analysis of these approaches.
Specifically, \method is built upon three key principles:
(1) \textit{Unification.} We provide a standardized interface that abstracts away model-specific details, enabling seamless integration and consistent evaluation across heterogeneous UMM architectures.
(2) \textit{Comprehensive coverage.}
\method supports a wide spectrum of multimodal capabilities, including understanding, generation, and editing, offering a holistic view.
(3) \textit{Flexibility.}
We introduce a modular mechanism to incorporate diverse post-training methods, allowing for easy application, research, and analysis.


Leveraging \method for large-scale, controlled evaluation across diverse models and tasks, we uncover several consistent and non-trivial findings.
First, current UMMs exhibit a highly fragmented capability landscape: no single model dominates across understanding, generation, and editing, and strong performance in one dimension often comes at the expense of others.
Second, model scale alone is \textit{not} a reliable predictor of performance; architectural design and training strategies play a more decisive role.
Third, we observe systematic trade-offs and partial decoupling between key capabilities, such as semantic correctness versus perceptual quality in editing, and generation quality versus multimodal reasoning.
Fourth, despite architectural unification, existing models struggle on tasks requiring tight coupling between perception, reasoning, and generation, particularly those involving structured manipulation or multi-step state tracking.
Finally, post-training methods are often unstable and model-dependent, frequently introducing cross-task degradation rather than holistic improvements.
Together, these results suggest that current UMMs achieve only limited unification, leaving true multimodal integration an open challenge.

\textbf{Contributions.}
This works makes the following contributions:
\begin{enumerate}[leftmargin=2em]
\setlength\itemsep{-.15em}
    \item \textbf{Foundational framework.} We present \method, the first comprehensive toolkit for UMMs that supports diverse models, datasets, and tasks. \method provides a foundational and flexible support for research and development of UMMs.
    \item \textbf{Comprehensive evaluation.} Based on \method, we conduct comprehensive evaluation studies, offering fair and reproducible evaluation reports for the community.
    \item \textbf{Insightful analysis.} Through extensive experiments, we demonstrate that widely used post-training techniques often introduce non-trivial performance trade-offs across tasks and benchmarks, highlighting the importance of holistic evaluation.
\end{enumerate}


\section{Related Work}
\label{sec-related}

\textbf{Unified Multimodal Models.}
Recent advances in unified multimodal models (UMMs) aim to integrate multimodal understanding and generation within a single architecture \citep{zhao2025unified, yin2024survey,deng2025emerging, chen2025blip3, geyer2023tokenflow, wang2026deepgen}. A dominant approach adopts decoder-only autoregressive transformers trained on interleaved multimodal tokens \citep{cui2025emu3, wang2024emu3}.
Another line of work explores hybrid generative frameworks that combine autoregressive modeling with diffusion or flow-based components \citep{xie2024show, wu2024janus, ma2024janusflow, chen2025janus}, which improve visual generation and cross-modal alignment. Additional efforts investigate modular or lightweight designs that bridge multimodal LLMs and generative models through intermediate connectors \citep{wu2025openuni}. Despite rapid progress, existing approaches primarily focus on architectural unification and scaling, leaving deeper issues of cross-modal consistency and reasoning less explored.

\textbf{Related Benchmarks and Codebase.}
Evaluating UMMs relies on a growing but fragmented ecosystem of benchmarks.
Benchmarks for multimodal understanding \citep{yue2024mmmu, fu2023mme, yu2023mm, liu2024mmbench, lu2023mathvista} assess reasoning, perception, and knowledge integration; generation benchmarks evaluate text-to-image fidelity, prompt alignment, and compositionality \citep{ghosh2023geneval, niu2025wise, li2026ueval, hu2024ella}; editing benchmarks focus on instruction-based image modification \citep{ye2025imgedit, liu2025step1x, zou2025uni}.
UMM-specific benchmarks \citep{wen2026unig2u, wu2026micon} further target cross-task generalization and multi-modal interaction within a single model.
Complementing these, open-source codebases \citep{deng2025emerging, wu2025omnigen2, wu2025openuni, dataflow2026openworldlib} provide training pipelines, pretrained models, while toolkits \citep{duan2024vlmevalkit, zhang2025lmms, contributors2023opencompass} offer standardized benchmarking interfaces.
However, their coverage remains partial and evaluation practices vary, with differences in preprocessing, protocols, and model interfaces, highlighting the need for a more unified frameworks.

\section{\method}

As shown in \Cref{fig:pipeline}, \method consists of four key layers: (1) \textit{Infrastructure layer}, which supports extension, common functions, and integration with other libraries; (2) \textit{Core functionality layer}, which implements the core functions including UMM base models, benchmark datasets, and post-training methods; (3) \textit{Task and execution layer}, which offers unifie pipelines for different tasks including understanding, generation, editing, and cross-task evaluation; and (4) \textit{Application and API layer}, which offers support for unified configuration, command line and Python interface, and evaluation reports.
Specifically, it currently supports $15$ models, $12$ benchmark datasets, and $5$ post-training methods.
Its modular design makes it easy to extend for more models, datasets, and training methods.

\vspace{-.1in}
\subsection{Supported Models}

TorchUMM supports a diverse set of UMM backbones under a standardized interface, enabling consistent evaluation across heterogeneous architectures.
It integrates a broad spectrum of models spanning different design paradigms and scales, including Bagel \citep{deng2025emerging}, OmniGen2 \citep{wu2025omnigen2}, Emu3 \citep{wang2024emu3}, Emu3.5 \citep{cui2025emu3}, the Janus series \citep{wu2024janus, ma2024janusflow, chen2025janus}, Show-o \citep{xie2024show}, Show-o2 \citep{xie2025show}, BLIP3-o \citep{chen2025blip3}, TokenFlow \citep{geyer2023tokenflow}, DeepGen \citep{wang2026deepgen}, MMaDA \citep{yang2025mmada}, and Ovis-U1 \citep{wang2025ovisu1}.

These models exhibit varying capabilities across three core multimodal functionalities: understanding, generation, and editing.
While most models support both understanding and generation, only a subset (e.g., Bagel, OmniGen2, DeepGen) natively supports image editing, enabling evaluation of more advanced multimodal manipulation behaviors.
By abstracting model-specific implementations into a unified interface, TorchUMM enables direct and fair comparisons across architectures without requiring changes to inference or evaluation pipelines.

\subsection{Supported Tasks and Datasets}

TorchUMM supports three primary tasks of UMMs spanning perception, creation, and manipulation:
\begin{itemize}[leftmargin=2em]
\setlength\itemsep{0em}
    \item \textbf{Multimodal Understanding}.
This task evaluates a model's ability to interpret and reason over visual inputs, including perception, visual question answering, and multimodal reasoning.
It covers both low-level perception and high-level cognitive understanding.

\item \textbf{Multimodal Generation}.
This task focuses on text-to-image generation, assessing a model's ability to synthesize images that align with textual prompts.
It evaluates aspects such as compositionality, semantic alignment, and incorporation of world knowledge.

\item \textbf{Multimodal Editing}.
This task evaluates a model's ability to modify existing images according to textual instructions.
It captures fine-grained control, consistency preservation, and instruction-following in editing scenarios.
\end{itemize}

More details of the corresponding benchmark datasets are in Appendix \ref{sec-append-torchumm-data}.

\subsection{Post-Training Methods}
\label{sec-method-post}

On top of evaluation on the three primary tasks, \method further supports unified post-training of UMMs.
Post-training is the fundamental step in adapting UMMs to downstream tasks to solve certain problems such as lack of consistency~\citep{su2025unigame} and for better generation~\citep{xie2025reconstruction}.
However, different post-training approaches often adopt different pipelines, implementation details, and benchmark datasets, making fair and reproducible comparison impossible.
\method supports several state-of-the-art post-training approaches through its unified interface, thus allowing for fair comparison.
These approaches are: (1) \textbf{SFT}: Supervised Fine-Tuning,
(2) \textbf{IRG}: Interleaving Reasoning for Generation \citep{huang2025interleaving},
(3) \textbf{UniCoT}: Unified Chain-of-Thought Reasoning \citep{qin2025uni},
(4) \textbf{RecA}: Reconstruction Alignment~\citep{xie2025reconstruction}, and
(5) \textbf{UniGame}: Self-play training~\citep{su2025unigame},
(6) \textbf{LatentUMM}: Dual Latent Alignment~\citep{luo2026latentumm},
(7) \textbf{UniPath}: Adaptive Coordination of Understanding and Generation~\citep{bai2026unipath}, and
(8) \textbf{FlexUMM}: Flexible Representation Learning~\citep{luo2026flexumm}.

\section{Evaluation Results}
\label{sec-benchmark}

We evaluated all supported models under a unified protocol using \method, ensuring identical preprocessing, inference settings, and scoring pipelines across architectures.
All results are independently reproduced; they do not represent official numbers from model authors.
This controlled setup enables, for the first time, direct and fair cross-architecture comparisons on generation, understanding, and editing tasks within a single framework.
All experiments were conducted on servers equipped with NVIDIA A100-80GB and H100-80GB GPUs, using multi-GPU parallelism where applicable to accommodate large-scale models and ensure efficient inference.

\subsection{Generation}
\label{sec-benchmark-gene}

\input{tab/generation-tab}
\Cref{tab:generation} summarizes text-to-image generation performance across three complementary benchmarks.
Several observations emerge from the unified evaluation.
Detailed results are in Appendix \ref{sec:detailed_results}.

First of all, \textbf{no single model dominates all generation facets}.
Ovis-U1 (3.6B) attains the highest GenEval score (90.05), surpassing DeepGen (86.59) and Emu3.5 (81.83) despite being the smallest 3.6B model in this comparison, and also achieves the strongest UEval result (34.17), indicating particularly compositional and instruction-faithful generation behavior.
DeepGen still leads on DPG Bench (87.44), excelling at fine-grained prompt fidelity, yet it lacks understanding capabilities entirely.
Emu3.5 leads on WISE (0.633) by a wide margin, suggesting superior encoding of world knowledge---particularly in cultural and spatial reasoning, but its generation pipeline differs substantially from other models, relying on vLLM \citep{kwon2023efficient}-based autoregressive decoding with classifier-free guidance rather than diffusion.
OmniGen2 and Bagel perform comparably on DPG Bench (84.51 vs.\ 84.11), indicating similar fidelity in detail-rich generation, while BLIP3-o achieves strong GenEval performance (81.36) despite its smaller 4B scale.
Second, \textbf{model scale alone does not determine generation quality}.
Ovis-U1 (3.6B) tops GenEval despite being smaller than most competitors; BLIP3-o (4B) outperforms the larger Janus-Pro (7B) on GenEval by 2.4 points and on WISE by 0.03, while the 1.3B Janus variants lag substantially behind all 7B models, suggesting that architectural design and training data quality play a more decisive role than parameter count.
Additionally, the WISE benchmark reveals a pronounced gap between models on world-knowledge-intensive prompts.
The spread from Emu3.5 (0.633) to Janus (0.222) is nearly threefold, indicating that knowledge grounding remains a key differentiator among current UMMs.

\subsection{Understanding}
\label{sec-benchmark-und}

\input{tab/understanding-tab}

\Cref{tab:understanding} presents multimodal understanding performance.
Models without visual understanding capability (DeepGen, BLIP3-o, TokenFlow) are excluded.
Key observations follow.

First, Bagel consistently dominates across both perception and reasoning benchmarks.
It achieves the best performance on all major metrics, including MME perception (1691.5), MME cognition (695.4), MMMU (0.519), MMBench-EN (0.8265), MMBench-CN (0.8094), MM-Vet (65.9), and MathVista (71.6).
Notably, its cognition score is more than twice that of most competitors, highlighting a substantial advantage in higher-order reasoning.
Among smaller-scale models, Ovis-U1 (3.6B) emerges as the strongest non-Bagel system on reasoning-heavy benchmarks, attaining MathVista 68.5 and MM-Vet 62.8---narrowly trailing the much larger Bagel (14B) and substantially outperforming similarly sized models such as OmniGen2 (63.5 / 62.7).
Second, \textbf{a clear perception–cognition gap persists across models}.
For example, Janus-Pro attains strong perception performance (1547.9) but much lower cognition (293.2), indicating difficulty in translating visual features into structured reasoning.
This gap is consistent across models and also reflected in cross-lingual evaluation, where performance on MMBench-CN typically lags behind MMBench-EN, suggesting additional challenges in multilingual multimodal understanding.
Finally, \textbf{understanding and generation exhibit a noticeable trade-off}.
Emu3.5, which performs strongly on generation benchmarks, achieves the lowest MME perception score (832.17) among models with understanding capability.
This suggests that architectures optimized for generation may compromise fine-grained visual understanding, whereas models like Bagel and OmniGen2 maintain a more balanced performance across both axes.

\subsection{Image Editing}
\label{sec-benchmark-edit}

\Cref{tab:editing} summarizes instruction-based image editing performance across GEdit-Bench (English and Chinese) and ImgEdit, covering both single-turn and multi-turn editing scenarios.

\input{tab/editing-tab}
First, Emu3.5 emerges as the strongest overall editing model.
It achieves the best overall scores on both GEdit-EN (7.56) and GEdit-CN (7.56), as well as the highest performance across all ImgEdit metrics, including SingleTurn (4.24), MultiTurn (4.89), and UGE (4.88).
Notably, Emu3.5 also attains the highest semantic correctness scores (7.64 EN, 7.62 CN), indicating strong capability in faithfully executing complex editing instructions while maintaining competitive perceptual quality.
Second, \textbf{a consistent semantic correctness-perceptual quality (SC--PQ) gap reveals a key limitation}.
OmniGen2 shows the largest disparity (e.g., 6.49 vs.\ 7.18 in English), indicating that it often preserves visual realism while failing to apply the intended semantic modification.
Bagel exhibits a smaller but still noticeable gap, while Emu3.5 and DeepGen maintain tighter alignment between SC and PQ.
This pattern suggests that \textit{semantic editing accuracy and perceptual quality preservation rely on partially decoupled capabilities}, and that current architectures tend to optimize the latter more reliably than the former.
Third, \textbf{multi-turn editing further amplifies model differences}.
On ImgEdit, Emu3.5 significantly outperforms all other models in both single-turn and multi-turn settings, with the largest margin observed in multi-turn editing (4.89 vs.\ 4.45 for Bagel, 3.27 for OmniGen2, and only 2.82 for Ovis-U1).
The contrast is particularly pronounced for Ovis-U1, which is competitive in single-turn editing (3.97) and on GEdit-EN overall (7.13), yet drops sharply once edits must be sustained over multiple turns---suggesting that single-turn editing strength does not automatically translate into reliable state tracking.
This indicates stronger consistency and state tracking across sequential edits, a critical requirement for real-world interactive editing systems.
Additionally, \textbf{cross-lingual performance is generally stable but model-dependent}.
DeepGen and Emu3.5 show consistent performance across English and Chinese, with only marginal variations, suggesting robust multilingual instruction following.
In contrast, OmniGen2 exhibits a noticeable drop in Chinese semantic correctness (6.49 $\to$ 6.25), indicating weaker cross-lingual alignment.
Bagel shows slight improvement in Chinese, potentially reflecting differences in training data composition.
Finally, \textbf{Editing performance highlights a broader generation--manipulation distinction}.
While several models achieve strong perceptual quality, only a subset (notably Emu3.5 and DeepGen) can reliably perform precise semantic modifications, especially under multi-step editing.
This suggests that instruction-based editing requires not only generative capacity but also fine-grained control and instruction grounding, which remain underdeveloped in many current UMMs.

\vspace{-.1in}
\subsection{Unification Evaluation}
\label{sec-benchmark-unified}

We evaluate the understanding--generation unification using Unified-Bench \citep{yan2025can}, RealUnified \citep{shi2025realunify}, and UEval \citep{li2026ueval}, which collectively assess whether models can form a consistent bidirectional loop between perception and generation in \Cref{tab:Unification-Benchmarks}.
It can be observed that leading UMMs begin to exhibit strong unification behavior, but it remains fragile and uneven across tasks.
On Unified-Bench, top models such as Bagel, Janus-Pro, and Omnigen2 achieve consistently high reconstruction scores, indicating that they can largely preserve semantic information through the I$\rightarrow$T$\rightarrow$I cycle.
Ovis-U1, despite its smaller size (3.6B), reaches a competitive overall score, suggesting that effective unification depends more on architectural design and training objectives than model scale.

\input{tab/unified-bench}

However, this apparent \textbf{unification is highly sensitive to the underlying visual representation space}.
Across nearly all models, performance drops significantly when moving from CLIP to structure-aware encoders such as DINOv2 and v3.
For example, Janus (1.3B) declines from 0.8465 (CLIP) to 0.5691 (DINOv3), and Show-o variants degrade even larger.
This gap indicates that while models can maintain coarse semantic alignment, they struggle to preserve fine-grained structural details, such as object relations and spatial layouts which are required for faithful reconstruction.
In contrast, LongCLIP consistently yields the highest scores, revealing a bias toward text-aligned embedding spaces that favor generation over structure-sensitive understanding.

RealUnify further exposes limitations in more realistic and compositional settings.
Even the best models achieve only moderate performance (e.g., 0.39 for Bagel and Ovis-U1), with noticeable variation across sub-tasks.
In particular, performance on compositional reasoning and multi-step inference remains limited, suggesting that current models lack robust mechanisms for maintaining consistency across intermediate reasoning steps.
UEval shows a similar trend, highlighting a gap between reconstruction-based consistency and task-level reasoning performance.
While some models perform well on Unified-Bench, their UEval results remain substantially lower (e.g., 30.9 for Bagel), indicating that preserving latent consistency does not necessarily translate to effective decision-making or problem-solving.

Finally, \textbf{model scale alone does not guarantee better unification}.
Despite having 34B parameters, Emu3.5 performs significantly worse (0.5093), reinforcing that unification is primarily governed by representation learning and training strategy rather than parameter count.
In summary, current UMMs partially succeed in aligning understanding and generation at a coarse semantic level, but fail to achieve true representation invariance across visual spaces and reasoning settings.
This suggest that explicit mechanisms for structure preservation, intermediate state tracking, and reasoning-aware generation are still missing, leaving tightly coupled unification tasks an open challenge.

\vspace{-10pt}
\subsection{Post-Training}
\vspace{-5pt}

\input{tab/posttrain-un-tab}
\input{tab/posttrain-edit-tab}
\input{tab/posttrain-gen-tab}
\Cref{tab:posttrain-generation,tab:posttrain-understanding,tab:posttrain-editing} systematically compare post-training strategies across multiple backbones and tasks. We analyze their effects across generation, understanding, and editing benchmarks.

First, \textbf{naive supervised fine-tuning (SFT) is insufficient to reliably improve unified multimodal models}.
While SFT yields marginal gains on certain metrics (e.g., MMMU: 0.519 $\rightarrow$ 0.526 on Bagel), it degrades performance across multiple benchmarks, including MMBench-EN/CN (0.8265/0.8094 $\rightarrow$ 0.8265/0.8088) and MM-Vet (65.9 $\rightarrow$ 61.2), as well as generation metrics such as DPG and WISE.
This trend is more pronounced on other backbones, where SFT causes substantial degradation (e.g., TokenFlow: DPG 71.29 $\rightarrow$ 22.16; Show-o2: consistent drops across DPG, GenEval, and WISE), suggesting that SFT tends to overfit to specific objectives while failing to preserve balanced multimodal capabilities.

Second, \textbf{the effectiveness of post-training is highly inconsistent across backbones and evaluation settings}.
The same method can yield different outcomes depending on the model.
For instance, SFT leads to mild degradation on Bagel but significantly larger drops on TokenFlow and Show-o2, while remaining relatively stable on OmniGen2.
Similarly, IRG introduces substantial and unstable degradation on Bagel across generation, understanding, and editing benchmarks, whereas RecA and UniCot exhibit more stable behavior, with modest improvements on MMMU and slight gains on MMBench-EN/CN.
These results indicate that post-training methods are sensitive to both backbone architecture and evaluation conditions, limiting their generalizability.

Finally, different multimodal capabilities exhibit heterogeneous sensitivity to post-training, leading to cross-task trade-offs.
Generation metrics are the most sensitive, where methods like UniGame improve compositional reasoning but degrade fidelity.
In contrast, understanding metrics remain relatively stable, including consistent performance across both MMBench-EN and MMBench-CN with only minor fluctuations.
Editing benchmarks show mixed and often marginal gains.

Overall, these patterns suggest that post-training tends to specialize models toward specific capabilities rather than uniformly improving all aspects, highlighting the challenge of achieving balanced multimodal improvement.

\vspace{-.1in}
\section{Analysis}
\label{sec-analysis}
\vspace{-.1in}
This section provides extensive analysis of UMM architectures for a more comprehensive understanding.
Architectural unification is often treated as a design prior for UMMs: sharing parameters, token spaces, and generation pathways should in principle encourage modalities and tasks to reinforce one another. We stress-test this assumption across MMaDA, Show-o2, and OmniGen2, whose coarse architectural unification degree follows MMaDA $>$ Show-o2 $>$ OmniGen2. The complete analysis is provided in Appendix \ref{sec-append-analysis}; here we preserve the main chain of evidence and summarize the central finding: \emph{architectural unification is not equivalent to representational unification}.

\noindent\textbf{Instruction following reverses the architectural prior.}
On UEval~\citep{li2026ueval}, models must interpret a text-to-image instruction and produce a visually grounded response. As shown in \Cref{fig:app-ueval-case-study}, the least architecturally unified model, OmniGen2, gives the strongest instruction-following behavior: it better preserves procedural structure in step-by-step drawing prompts and more often produces recognizable technical layouts for diagram prompts. Show-o2 tends to produce visually plausible but semantically under-specified images, while MMaDA more often collapses in both grounding and visual organization. This does not by itself identify a clean causal effect of unification, since the models also differ in backbone, generator, optimization recipe, and data. It does show, however, that the practical capability ordering can be nearly the reverse of the architectural unification prior.

\begin{figure}[t!]
\centering
\setlength{\tabcolsep}{3pt}
\begin{tabular}{ccc}
\textbf{MMaDA-8B} & \textbf{Show-o2-7B} & \textbf{OmniGen2-8B} \\
\includegraphics[width=0.315\textwidth]{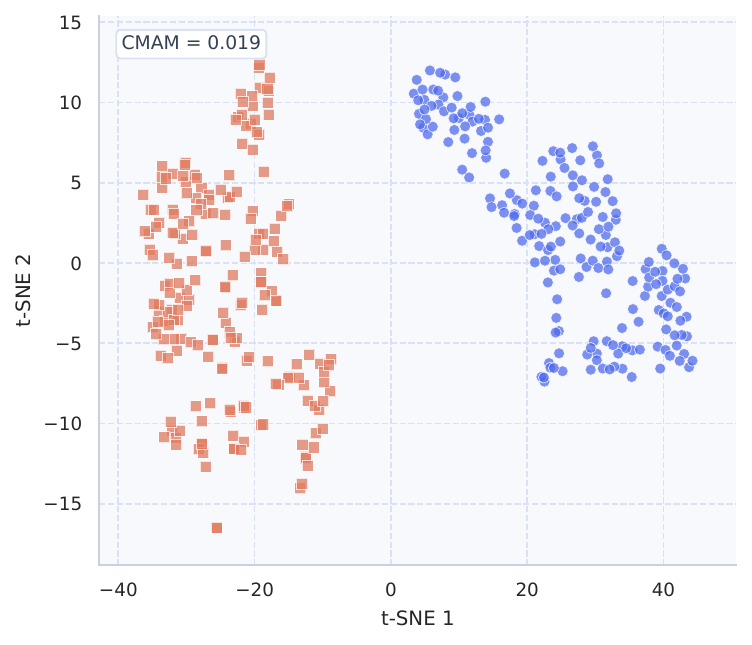} &
\includegraphics[width=0.315\textwidth]{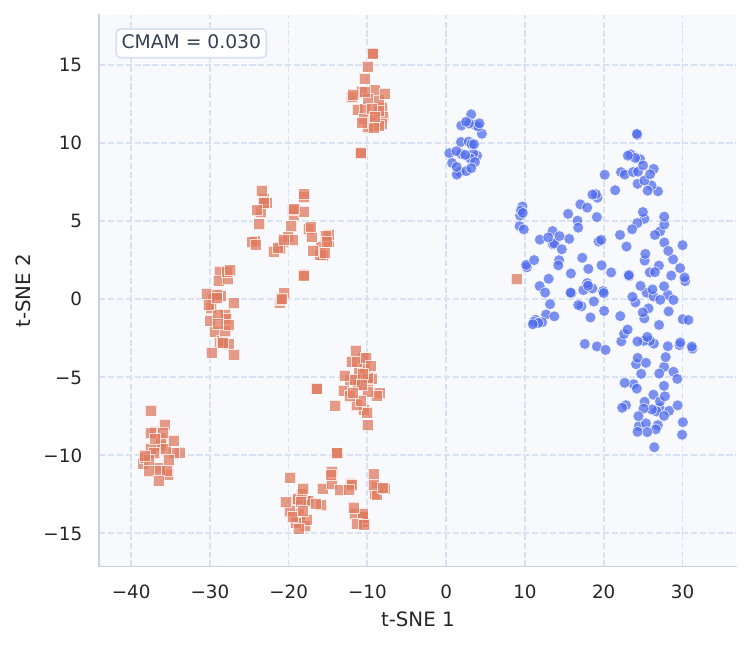} &
\includegraphics[width=0.315\textwidth]{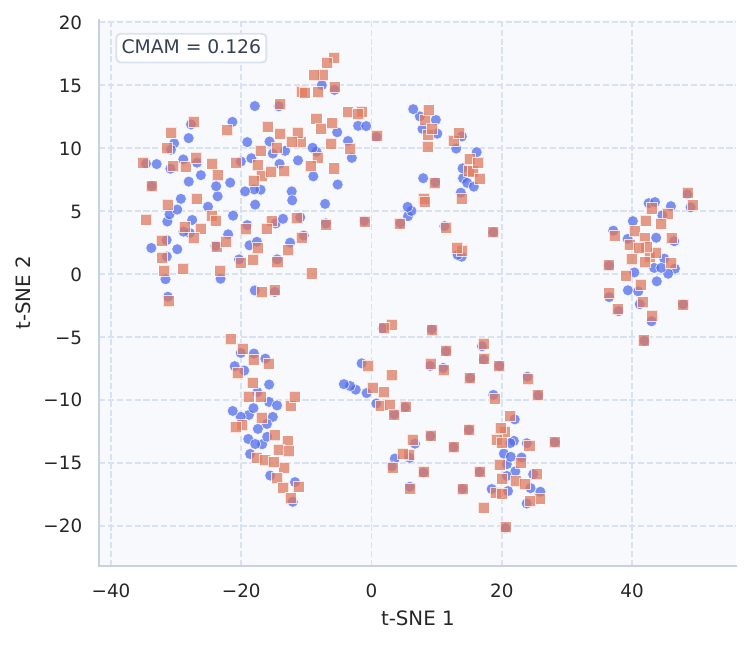}
\end{tabular}\\[-0.3em]
\includegraphics[width=0.28\textwidth]{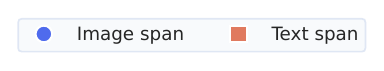}
\vspace{-.1in}
\caption{Final-layer UEval latent spaces. CMAM measures how much closer a matched text--image pair is than mismatched pairs; larger values indicate a smaller effective modality gap. OmniGen2, although least unified by design, has the strongest cross-modal alignment.}
\label{fig:analysis-latent-gap}
\vspace{-.2in}
\end{figure}

\noindent\textbf{Unification training can destabilize inherited behavior.}
We then ask whether unification preserves or rewrites the behavior of the backbone from which a model starts. On a shared 200-sample MathVista subset, we generate meaning-preserving query variants and compare both external response embeddings and internal response-span representations across layers. \Cref{fig:app-analysis-query-consistency} shows that OmniGen2 remains close to its Qwen2.5-VL-3B backbone: response consistency and layerwise latent dynamics are nearly unchanged. Show-o2, by contrast, deviates substantially from its Qwen2.5-VL-7B initializer, with broader response-embedding dispersion and lower internal consistency through the middle layers. Thus, stronger sharing across understanding and generation does not merely add capability; it can also rewrite useful inherited properties.

\begin{figure}[!t]
\centering
\setlength{\tabcolsep}{3pt}
\begin{tabular}{ccc}
\includegraphics[width=0.31\textwidth]{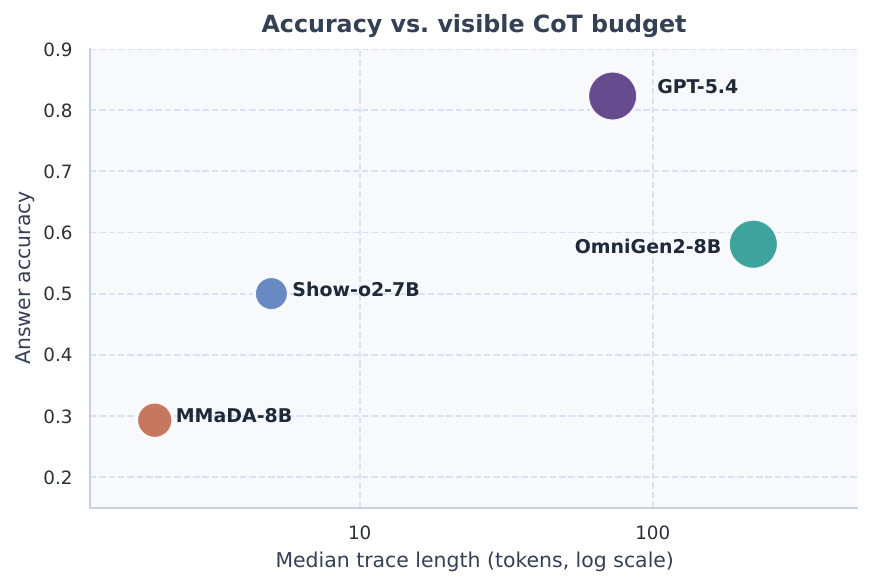} &
\includegraphics[width=0.31\textwidth]{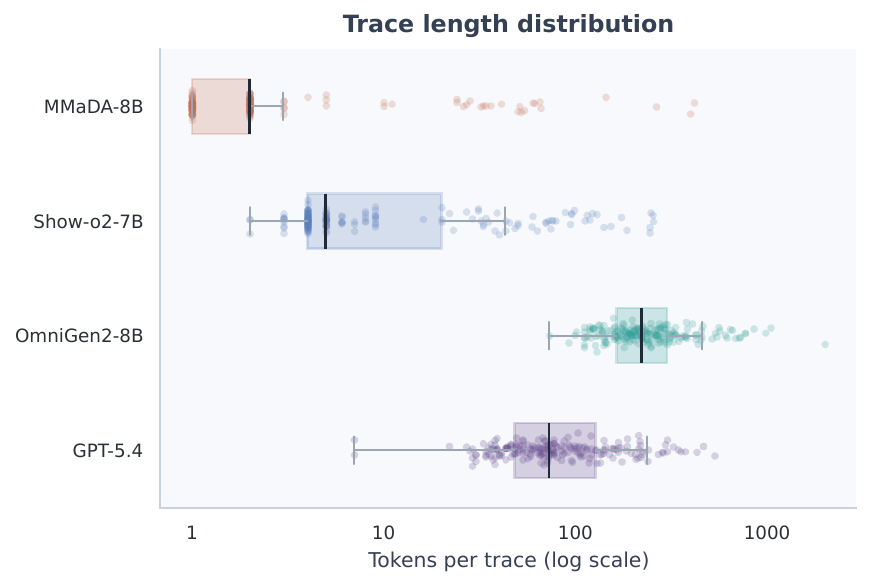} &
\includegraphics[width=0.31\textwidth]{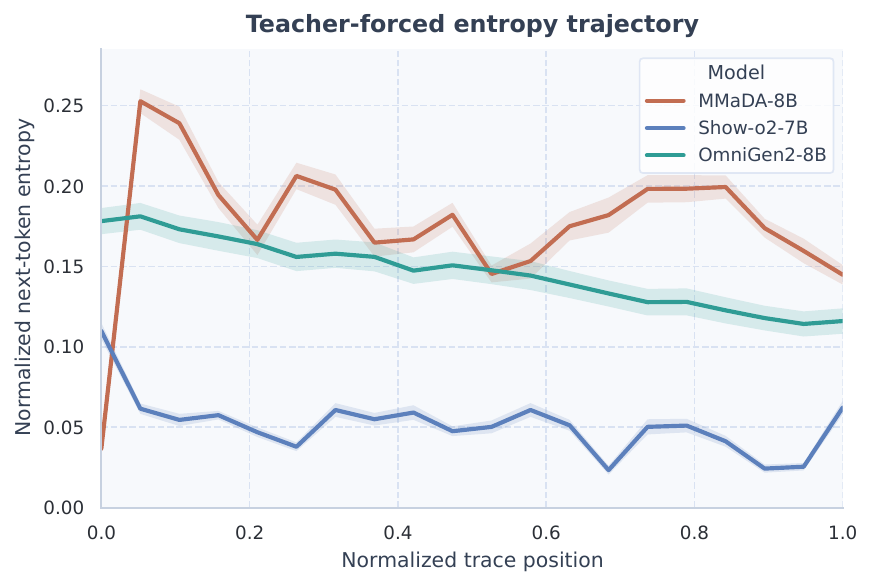}
\end{tabular}
\vspace{-.1in}
\caption{Reasoning-trace behavior on MathVista. From left to right: answer accuracy against median visible trace length; the full trace-length distribution; and teacher-forced entropy trajectories for the three open-source UMMs. OmniGen2, the least architecturally unified model among the three, most reliably follows the CoT instruction, produces long explicit traces, and achieves the best open-source accuracy. MMaDA and Show-o2, which apply more aggressive sharing across modalities and tasks, frequently collapse to short responses and fail to obtain comparable test-time scaling.}
\label{fig:app-analysis-reasoning}
\vspace{-.2in}
\end{figure}

\noindent\textbf{CoT traces expose a test-time scaling gap.}
Reasoning gives a third view of the same trade-off. Under explicit chain-of-thought prompting on MathVista, \Cref{fig:app-analysis-reasoning} shows that OmniGen2 produces visible CoT traces for almost all samples (99.5\%), with a median length of 220 tokens and the best open-source accuracy (58.1\%). MMaDA and Show-o2 much more often collapse to short answer-style responses: their median visible lengths are only 2 and 5 tokens, and answer-only responses appear in 63.1\% and 49.5\% of samples, respectively. Their accuracies, 29.3\% and 50.0\%, lag behind OmniGen2 despite higher architectural unification. The failure mode is therefore not only lower final accuracy, but an inability to convert a CoT instruction into useful test-time computation.

\noindent\textbf{The common signature is the modality gap.}
The behavioral probes above point to a representational explanation. If architectural unification were successfully producing a shared multimodal space, paired text and image states should become closer inside the model. We test this on UEval by re-encoding each model's own \((\text{prompt}, \text{image answer}, \text{text answer})\) trace through its multimodal understanding pathway, pooling the text-answer span and image-answer span at the final layer, and comparing matched versus mismatched text--image pairs. For L2-normalized text and image pooled features $\hat{z}^{\ell}_{t,i}$ and $\hat{z}^{\ell}_{v,i}$, we define Cross-Modal Alignment Margin (CMAM) as
\begin{equation*}
\mathrm{CMAM}^{\ell}
= \frac{1}{N}\sum_{i=1}^{N}
\cos\!\left(\hat{z}^{\ell}_{t,i}, \hat{z}^{\ell}_{v,i}\right)
- \frac{1}{N(N-1)}\sum_{i\neq j}
\cos\!\left(\hat{z}^{\ell}_{t,i}, \hat{z}^{\ell}_{v,j}\right).
\end{equation*}
Larger CMAM indicates that paired modalities are more tightly aligned, i.e., a smaller effective modality gap.
\Cref{fig:analysis-latent-gap} shows that the latent geometry again reverses the architectural prior. OmniGen2 has the largest CMAM (0.126) and visibly interleaves paired text and image spans, while Show-o2 and MMaDA have much smaller margins (0.030 and 0.019). This suggests that more aggressive cross-modal and cross-task parameter sharing can widen, rather than close, the practical modality gap before sufficient scale, data, or optimization support is reached. The implication is not that unification is undesirable; rather, future UMM development should distinguish \emph{architectural} unification from the harder goal of obtaining a genuinely aligned multimodal representation space.

\vspace{-5pt}
\section{Conclusion and Discussion}
\label{sec-conclusion}
\vspace{-5pt}

We present \method, a unified framework for evaluating, analyzing, and post-training UMMs across diverse models, benchmarks, and training methods.
By standardizing inference and scoring under a shared pipeline, \method enables reliable comparisons that are difficult to obtain from previously fragmented evaluation setups.

Several insights emerge only under this controlled setting.
We observe a consistent gap between perception and higher-level cognition in multimodal understanding, as well as a mismatch between perceptual quality and semantic correctness in image editing tasks.
While current models perform reasonably well on static understanding and loosely grounded reasoning, they struggle with tasks requiring iterative and structured visual manipulation, which becomes apparent only when understanding and generation are evaluated jointly.
Our post-training analysis further reveals that the same training method can behave inconsistently across different backbones, sometimes improving one model while degrading another, highlighting strong backbone dependence.
In addition, we find that more tightly unified architectures do not necessarily guarantee better compositional reasoning, and may even destabilize certain inherited capabilities.

Overall, these findings suggest that progress in UMMs is highly sensitive to evaluation protocol, model architecture, and training setup, and that conclusions drawn from isolated benchmarks or single-model studies can be misleading.

\textbf{Limitations.}
Our evaluation is restricted to publicly released checkpoints under standard settings, and alternative prompting or decoding strategies may lead to different outcomes.
The benchmark suite is still largely focused on English and static images, leaving modalities such as video, audio, and human preference modeling for future work.
Finally, while our post-training study covers a broader range of methods than prior work, it still samples only a limited portion of the full design space across models, methods, and data.

\bibliographystyle{plain}
\bibliography{refs}

\appendix

\input{appendix}

\newpage
\input{checklist.tex}

\end{document}

%% file: tab/generation-tab.tex
\begin{table}[t]
\centering
\caption{Text-to-image generation results. DPG Bench and GenEval are reported as overall accuracy (\%); WISE reports the WiScore (0--1) evaluated by Qwen2.5-VL-72B-Instruct.} 
\vspace{-.1in}
\label{tab:generation}
\resizebox{.7\textwidth}{!}{
\begin{tabular}{lccccc}
\toprule
Model & Venue & Architecture & DPG Bench  & GenEval  & WISE   \\
\midrule
Show-o (1.3B)  & ICLR'25 & AR+diffusion & 78.74 & 65.06 & 0.304 \\
Emu3 (8B)    &  arxiv2409 & AR & 80.31 & 45.76 & 0.337\\
Janus (1.3B)   & arxiv2410 & AR &  73.53 & 40.04 & 0.222 \\
JanusFlow (1.3B) & CVPR'25 & AR + diffusion &  72.03 & 49.99 & 0.296 \\
TokenFlow (7B)  & CVPR'25 & AR & 71.29 & 52.21 & 0.306 \\
Janus-Pro (7B) & arxiv2501 & AR & 83.73 & 78.92 & 0.381 \\
BLIP3-o (4B)  & arxiv2505 & AR+diffusion & 61.47 & 81.36 & 0.414 \\
Bagel (14B)    &  arxiv2505 &  AR+diffusion & 84.11 & 78.81 & 0.399  \\
MMaDA (8B)    & NeurIPS'25  & diffusion & 64.55 & 46.12 & \textbf{0.656} \\
Show-o2 (1.5B)  & arxiv2506 & AR & 82.78 & 55.49 & 0.335  \\
Show-o2 (7B)  & arxiv2506 & AR & 82.81 & 59.87 & 0.360  \\
OmniGen2 (7B)  & arxiv2506 &  AR+diffusion & 84.51 & 78.53 & 0.403  \\
Ovis-U1 (3.6B) & arxiv2506 & AR+diffusion & 62.21 & \textbf{90.05} & 0.376  \\
Emu3.5 (34B)  & arxiv2510 & AR &    72.51  &   81.83   & 0.633\\
DeepGen (5B)   & arxiv2602 &  AR+diffusion &   \textbf{87.44}   & 86.59 & 0.547 \\
\bottomrule
\end{tabular}}
\vspace{-.1in}
\end{table}

%% file: tab/understanding-tab.tex
\begin{table}[t]
\centering
\caption{Multimodal understanding results. MME reports perception (P) and cognition (C) scores.  
}
\vspace{-.1in}
\label{tab:understanding}
\resizebox{.8\textwidth}{!}{
\begin{tabular}{lcccccccc}
\toprule
Model & MME (P) & MME (C) & MMMU & MMBench(EN) & MMBench(CN) & MM-Vet & MathVista & MMStar \\
\midrule
Bagel (14B)       & \textbf{1691.5} & \textbf{695.4} & \textbf{0.519} & \textbf{0.8265} & \textbf{0.8094} & \textbf{65.9} & \textbf{71.6} & \textbf{63.20} \\
Janus-Pro (7B)   & 1547.9 & 293.2 & 0.407 & 0.6764 & 0.6494 & 33.7 & 42.8 & 42.00 \\
JanusFlow (1.3B) & 1305.6 & 251.1 & 0.290 & 0.6416 & 0.6031  & 31.8 & 34.8 & 39.19 \\
Janus (1.3B)     & 1221.4 & 264.3 & 0.273 & 0.2769 & 0.3719  & 27.0 & 26.6 & 26.67 \\
Show-o2 (7B)     & 1619.8 & 387.5 & 0.479 & 0.7725 & 0.7668 & 47.1 & 51.5 & 54.53 \\
Show-o2 (1.5B)   & 1413.3 & 291.8 & 0.371 & 0.6509 & 0.6104 & 46.1 & 37.9 & 43.87 \\
Show-o (1.3B)    & 1188.5 & 244.6 & 0.261 & 0.4348 & 0.1101 & 23.3 & 29.0 & 27.87 \\
Emu3 (8B)        & 1176.0 & 213.2 & 0.314 & 0.5953 & 0.4566  & 30.0 & 44.9 & 43.60 \\
Emu3.5 (34B)     &  832.2 & 271.4 & 0.292 & 0.1756 & 0.1813  & 28.0 & 30.6 & 30.27 \\
OmniGen2 (7B)    & 1584.4  & 614.6 & 0.460 & 0.7636 & 0.7543  & 62.7 & 63.5 & 52.47 \\
Ovis-U1 (3.6B)   & 1547.6 & 575.4 & 0.437 & 0.7855 & 0.7699 & 62.8 & 68.5 & 58.27 \\
MMaDA (8B)       & 939.0 & 241.4 & 0.289 & 0.2816 & 0.2057 & 11.4 & 24.9 & 31.73 \\
\bottomrule
\end{tabular}
}
\vspace{-.2in}
\end{table}

%% file: tab/editing-tab.tex
\begin{wraptable}{r}{0.55\textwidth}
\vspace{-.25in}
\centering
\caption{Image editing results. SC = Semantic Correctness, PQ = Perceptual Quality, O = Overall (geometric mean), ST = SingleTurn, MT = MultiTurn, all on a 0--10 scale. Evaluated by Qwen2.5-VL-72B via VIEScore.}
\label{tab:editing}
\resizebox{\linewidth}{!}{
\begin{tabular}{lccccccccc}
\toprule
 & \multicolumn{3}{c}{GEdit-EN} & \multicolumn{3}{c}{GEdit-CN} & \multicolumn{3}{c}{ImgEdit} \\
\cmidrule(lr){2-4} \cmidrule(lr){5-7}  \cmidrule(lr){8-10}
Model & SC & PQ & O & SC & PQ & O & ST & MT & UGE \\
\midrule
DeepGen  & 7.44 & \textbf{7.54} & 7.33 & 7.41 & \textbf{7.59} & 7.36 & 4.07 & 4.37 & 4.81 \\
Bagel    & 6.68 & 7.04 & 6.35 & 6.83 & 7.06 & 6.52 & 3.71 & 4.45 & 4.18  \\
OmniGen2 & 6.49 & 7.18 & 6.27 & 6.25 & 7.18 & 6.03 & 3.88 & 3.27 & 4.06  \\
Emu3.5   & \textbf{7.64} & 7.48 & \textbf{7.56} & \textbf{7.62} & 7.50 & \textbf{7.56} & \textbf{4.24} & \textbf{4.89} & \textbf{4.88}  \\
Ovis-U1  & 7.28 & 7.46 & 7.13 & 6.78 & 7.51 & 6.72 & 3.97 & 2.82 & 4.48 \\
\bottomrule
\end{tabular}
}
\vspace{-10pt}
\end{wraptable}

%% file: tab/unified-bench.tex
\begin{table}[t]
\centering
\vspace{-15pt}
\caption{Results on unification benchmarks.}
\vspace{-.05in}
\label{tab:Unification-Benchmarks}
\resizebox{.9\textwidth}{!}{
\begin{tabular}{l|ccccc|ccccc|c}
\toprule
\multirow{2}{*}{Model} 
& \multicolumn{5}{c|}{Unified-Bench} 
& \multicolumn{5}{c|}{RealUnified~(GEU)} 
& \multirow{2}{*}{UEval} \\
 & Clip & Dinov2 & Dinov3 & Longclip & Overall & MC & CN & AF & MR & Total \\ \midrule
Bagel~(14B) & 0.8947 & 0.7877 & 0.7240 & 0.9321 & 0.8346 & 0.30 & 0.39 & 0.52 & 0.34 & 0.39 & 30.9 \\
Janus-Pro~(7B) & 0.8874 & 0.7753 & 0.7232 & 0.9312 & 0.8293 & 0.20& 0.26 & 0.31 & 0.26 & 0.26 & 20.6 \\
Janus~(1.3B) & 0.8465 & 0.6396 & 0.5691 & 0.8982 & 0.7384 & 0.04 & 0.24 & 0.17 & 0.20 & 0.16 & -- \\
Janus-flow~(1.3B) & 0.8290 & 0.5598 & 0.4775 & 0.8734 & 0.6849 & 0.24 & 0.20 & 0.26 & 0.22 & 0.23 & -- \\
Show-o2~(7B) & 0.8281 & 0.5859 & 0.4994 & 0.8679 & 0.6953 & 0.26 & 0.30 & 0.38 & 0.31 & 0.31 & 29.1 \\
Show-o2~(1.5B) & 0.7980 & 0.4761 & 0.3895 & 0.8304 & 0.6235 & 0.20 & 0.29 & 0.27 & 0.28 & 0.26 & 22.1 \\
Show-o~(1.3B) & 0.7682 & 0.4446 & 0.4288 & 0.8197 & 0.6153 & 0.21 & 0.28 & 0.28 & 0.23 & 0.25 & -- \\
Omnigen2~(7B) & 0.8894 & 0.7598 & 0.7283 & 0.9307 & 0.8271 & 0.25 & 0.26 & 0.54 & 0.31 & 0.34 & 25.8 \\
Emu3.5~(34B) & 0.6807 & 0.3239 & 0.3295 & 0.7030 & 0.5093 & 0.28 & 0.25 & 0.31 & 0.36 & 0.30 & -- \\
MMaDA~(8B) & 0.6750 & 0.2492 & 0.2310 & 0.7064 & 0.4654 & 0.23 & 0.21 & 0.30 & 0.24 & 0.30 & 4.4\\
Ovis-u1~(3.6B) & 0.8671 & 0.7498 & 0.6721 & 0.9312 & 0.8051 & 0.29 & 0.36 & 0.60 & 0.31 & 0.39 & 34.2 \\ \bottomrule
\end{tabular}
}
\vspace{-15pt}
\end{table}

%% file: tab/posttrain-un-tab.tex
\begin{table}[t]
\centering
\caption{Post-training results on understanding benchmark.
}
\vspace{-.1in}
\label{tab:posttrain-understanding}
\resizebox{.8\textwidth}{!}{
\begin{tabular}{lccccccc}
\toprule
Variant  & MME (P) & MME (C) & MMMU & MMBench(EN) & MMBench(CN) & MM-Vet & Mathvista \\
\midrule
Bagel (base)      & 1691.5 & 695.4 & 0.519 & 0.8265 & 0.8094 & 65.9 & 71.6 \\
\quad + RecA      & 1689.1 & 695.4 & 0.523 & 0.8265 & 0.8094 & 66.1  & 51.6 \\
\quad + UniCot    & 1690.7 & 678.2 & 0.531 & 0.8312 & 0.8099 & 64.5 & 73.0 \\
\quad + SFT       & 1680.7 & 678.9 & 0.526 & 0.8265 & 0.8088 & 61.2 & 73.1 \\
\quad + IRG       & 1647.5 & 650.4 & 0.480 & 0.6047 & 0.5735 & 40.7 & 68.0 \\
\quad + UniGame   & 1692.1 & 695.4 & 0.524 & 0.8275 & 0.8094 & 60.7 & 72.2 \\
\quad + LatentUMM & 1695.8 & 682.6 & 0.541 & 0.8499 & 0.8491 & 67.2 & 73.6 \\
\quad + Unipath & 1669.4 & 694.64 & 0.541 & 0.8631 & 0.8357 & 50.6 & 72.2 \\
\quad + FlexUMM & 1685.3 & 688.7 & 0.540 & 0.8675 & 0.8587 & 66.1 & 73.4\\
Janus-Pro + UniGame   & 1554.0 & 288.9 & 0.409 & 0.6732 & 0.6519 & 32.4 & 43.9 \\
Janus-Pro + SFT   & 1549.9 & 292.9 & 0.400 &  0.6748 & 0.6514 & 33.0 & 35.4 \\
Omnigen2 + SFT   & 1573.6 & 610.0 & 0.469 & 0.7636 & 0.7543 & 62.2 & 63.5 \\
\bottomrule
\end{tabular}
}
\vspace{-.1in}
\end{table}

%% file: tab/posttrain-edit-tab.tex
\begin{table}[t]
\centering
\caption{Post-training on editing benchmarks. For GEdit-EN and GEdit-CN, (I/F) denotes Intersection and Full scores, respectively. ImgEdit results are reported as S/M/U (single-turn, multi-turn, UGE).
}
\vspace{-.1in}
\label{tab:posttrain-editing}
\resizebox{.8\textwidth}{!}{
\begin{tabular}{lccccc}
\toprule
Variant & GEdit-EN(I/F) & GEdit-CN(I/F) & ImgEdit(S) & ImgEdit(M) & ImgEdit(U) \\
\midrule
Bagel (base)      & 6.38/6.35 & 6.68/6.52 & 3.71 & 4.45 & 4.18 \\
\quad + RecA       & 6.89/6.80 & 6.87/6.75 & 3.89 & 4.28 & 4.15 \\
\quad + UniCot     & 7.04/6.92 & 6.90/6.81 & 3.77 & 4.22 & 4.34  \\
\quad + SFT         & 6.62/6.49 & 6.71/6.54 & 3.73 & 4.48 & 4.12 \\
\quad + IRG        & 6.52/6.44 & 6.51/6.41 & 3.79 & 3.89 & 4.54\\
\quad + UniGame     &  6.48/6.48 & 6.55/6.38 & 3.72 & 4.46 & 4.31 \\
Omnigen2 + SFT      & 6.63/6.56 & 6.54/6.43 & 3.73 & 3.19 & 4.04 \\
\bottomrule
\end{tabular}

}
\vspace{-.2in}
\end{table}

%% file: tab/posttrain-gen-tab.tex

\begin{wraptable}{r}{0.48\textwidth}
\vspace{-20pt}
\centering
\caption{Post-training results for generation.}
\label{tab:posttrain-generation}
\resizebox{.5\textwidth}{!}{
\begin{tabular}{lcccc}
\toprule
Variant & DPG & GenEval & WISE & UEval  \\
\midrule
Bagel (base)      & 84.11 & 78.81 & 0.399 & 30.9 \\
\quad + RecA       &  \textbf{85.20} & 83.05 & \textbf{0.423} & 31.0\\
\quad + UniCot     &  83.52    &   77.91   &  0.404  &  31.8   \\
\quad + SFT        &  83.02 & 78.03 &   0.227   & \textbf{31.4} \\
\quad + IRG        &   81.82   & 72.06 & 0.384 & 9.1\\
\quad + UniGame        &   65.77   & \textbf{85.80} & 0.403 &  31.0 \\
Janus-Pro + UniGame        &   83.92   & 78.65 & 0.373 &  20.7 \\
Janus-Pro + SFT        &   83.93   & 77.61 & 0.370 &  20.6 \\
Omnigen2 + SFT        &   83.76   & 77.23 & 0.392 &  20.6 \\
Blip3-o + SFT        &   61.01   & 78.41 & 0.399 &  N/A \\
Tokenflow + SFT        &   22.16   & 51.96 & 0.328 &  N/A \\
Show-o2(7B) + SFT        &   80.58   & 52.13 & 0.322 &  25.7 \\
\bottomrule
\end{tabular}
}
\vspace{-10pt}
\end{wraptable}

%% file: appendix.tex
\onecolumn

\begin{center}
    {\large \textbf{Appendix\\
    \method: A Unified Multimodal Model Codebase for Evaluation, Analysis, and Post-training}}
\end{center}

\etocdepthtag.toc{mtappendix}
\etocsettagdepth{mtchapter}{none}
\etocsettagdepth{mtappendix}{subsection}
\tableofcontents

\section{Codebase Structure of \method}
\label{sec-codebase}
TorchUMM is designed as a modular framework that minimizes the engineering effort required to integrate new models, benchmarks, and post-training methods.
The codebase is organized around four principal components: backbone adapters, a configuration system, running pipelines, and cloud infrastructure.

\subsection{Backbone Adapter Architecture}
The core abstraction in TorchUMM is the \texttt{BackboneAdapter} protocol which defines a minimal interface that every model must implement:

\begin{tcblisting}{
  colback=gray!10,
  colframe=gray!50,
  width=\columnwidth,
  listing only,
  title=Adapter Architecture
}
class BackboneAdapter(Protocol):
    name: str
    def load(self, cfg: dict) -> None: ...
    def generate(self, batch: dict, gen_cfg: dict) -> Any: ...
\end{tcblisting}

Each supported model is wrapped in a self-contained adapter module under \texttt{src/umm/backbones/} that handles model-specific concerns, including device mapping, tokenization, attention implementation and output formatting behind the standardized interface.

\subsection{Configuration System}
All behavior in TorchUMM is driven by YAML configuration files organized into three layers:

\begin{itemize}[leftmargin=2em]
\setlength\itemsep{0em}
    \item \textbf{Inference configs} specify per-model settings: model weights path, generation hyperparameters (e.g., number of diffusion timesteps, classifier-free guidance scales), and device allocation.
    \item \textbf{Evaluation configs} bind a backbone to a benchmark, specifying dataset paths, output directories, evaluation mode (single-stage or two-stage), and scoring model selection.
    \item \textbf{Post-training configs} define training pipelines, including the method (SFT, recA, IRG, UniCot), optimizer settings, checkpoint intervals, and distributed training parameters.
\end{itemize}

Switching from one model to another on the same benchmark requires changing only the backbone name and model path in the config file; no code modifications are necessary.
This separation of concerns enables rapid experimentation and ensures reproducibility, as each evaluation run is fully determined by its config file.

\subsection{Running Pipeline of \method}
The pipeline is organized into three operational stages: inference, evaluation, and post-training, each driven entirely by YAML configs and a shared backbone adapter interface.

\subsubsection{Inference Pipeline}
The inference stage instantiates an InferencePipeline from an inference config, which specifies the backbone name, model weights, and generation/editing/understanding parameters. The pipeline normalizes each request into a unified Inference Request (prompt, images, task, and params), then routes it to the task-specific handler (generation, understanding or editing).
Switching models requires only changing \texttt{inference.backbone} and \texttt{inference.backbone\_cfg} in the YAML config.

\subsubsection{Evaluation Pipeline}
Evaluation runs through the \texttt{umm eval} CLI, which loads a YAML config, dispatches to a benchmark-specific runner based on \texttt{eval.benchmark}, and then executes dataset iteration and scoring. Each benchmark wrapper builds an \texttt{InferencePipeline}, formats prompts and inputs, and saves structured outputs (JSON/JSONL/XLSX) for analysis or external scoring scripts.
Two-stage benchmarks (e.g., GenEval, UEval, WISE, DPG-Bench, GEdit/ImgEdit) are supported via thin wrappers that call official scripts while preserving TorchUMM's standard config, logging, and output layout.

\subsubsection{Post-Training Pipeline}
Post-training is configured through YAML files that specify the method (SFT, recA, IRG, UniCot), optimizer settings, checkpoint schedule, and distributed training parameters. Training entry points live under \texttt{src/umm/post\_training/}, and the resulting checkpoints can be evaluated immediately by pointing an evaluation config to the new model path.
This design keeps training logic isolated from evaluation logic, enabling rapid iteration without changing core inference or benchmark code.

\subsection{Extensibility}
\label{sec-code-exten}

Adding a new component to TorchUMM follows a minimal-touch pattern:
\begin{itemize}[leftmargin=2em]
\setlength\itemsep{0em}
    \item \textbf{New model:} Implement a \texttt{BackboneAdapter} subclass, register it in the backbone registry, and create inference/evaluation config files. No changes to evaluation scripts or infrastructure code are needed.
    \item \textbf{New benchmark:} Implement a benchmark handler (generation and/or scoring scripts), add a CLI route, and create config files. The existing \texttt{InferencePipeline} handles all model interaction.
    \item \textbf{New post-training method:} Implement the training loop under \texttt{src/umm/post\_training/}, add a config, and register the pipeline entry point. The trained model can then be evaluated using existing configs by simply changing the model weights path.
\end{itemize}

\section{Details of \method}
\label{sec-append-torchumm}

\subsection{Supported Datasets and Benchmarks}
\label{sec-append-torchumm-data}


\textbf{Generation Benchmarks.}
We include representative text-to-image evaluation benchmarks: (1) \textbf{DPG-Bench} \citep{hu2024ella}, which focuses on fine-grained detail preservation and prompt fidelity by evaluating whether generated images accurately reflect complex attributes and subtle visual concepts described in text; (2) \textbf{GenEval} \citep{liu2025step1x}, which emphasizes compositional generalization by assessing a model's ability to correctly combine multiple objects, attributes, and relations within a single scene; (3) \textbf{WISE} \citep{niu2025wise}, which evaluates world knowledge integration and semantic correctness, requiring models to generate images that align with commonsense and factual knowledge beyond the literal prompt.

\textbf{Understanding Benchmarks.}
For multimodal understanding, we adopt widely used benchmarks including (1) \textbf{MMMU} \citep{yue2024mmmu}, which evaluates expert-level reasoning across diverse academic disciplines by requiring models to integrate visual perception with domain-specific knowledge. (2) \textbf{MMBench} \citep{liu2024mmbench}, which provides a comprehensive assessment of general multimodal understanding through carefully designed multiple-choice questions covering perception, reasoning, and instruction following. (3) \textbf{MME} \citep{fu2023mme}, which focuses on disentangling perception and cognition abilities via a suite of diagnostic tasks targeting object recognition, OCR, commonsense reasoning, etc. (4) \textbf{MM-Vet} \citep{yu2023mm}, which emphasizes complex reasoning and robustness through open-ended questions that require multi-step inference and cross-modal grounding and  (5) \textbf{MathVista} \citep{lu2023mathvista}, which evaluates mathematically grounded visual reasoning, requiring models to interpret diagrams, charts, and geometric figures to solve quantitative problems. (6) \textbf{MMStar} \citep{chen2024mmstar}, which evaluates vision-dependent multimodal reasoning through a curated set of challenging samples designed to minimize data leakage and assess fine-grained multimodal capabilities.

\textbf{Unified and Cross-Task Benchmarks.}
To better align with the capabilities of UMMs, we include benchmarks specifically designed for integrated and cross-task evaluation, including (1) \textbf{UEval}~\citep{li2026ueval}, which provides a unified evaluation framework that spans diverse generation and reasoning scenarios, enabling holistic assessment across multiple capabilities.
(2) \textbf{Uni-MMMU}~\citep{zou2025uni}, which extends traditional multimodal evaluation by jointly assessing understanding, generation, and editing within a single benchmark, emphasizing cross-task generalization and consistency.
These benchmarks move beyond isolated task evaluation and instead measure how well models perform under realistic settings that require coordinating multiple capabilities within a unified framework.
(3) \textbf{RealUnify}~\citep{shi2025realunify}, a benchmark designed to evaluate whether unified multimodal models can achieve bidirectional capability synergy between understanding and generation. It includes human-annotated tasks structured along two complementary axes: understanding enhances generation, where reasoning is required to guide image synthesis, and generation enhances understanding, where generative simulation supports problem solving. 

\textbf{Editing Benchmarks.}
For image editing, we incorporate benchmarks including (1) \textbf{GEdit-Bench} \citep{liu2025step1x}, which evaluates instruction-based image editing with a focus on precise attribute manipulation, multi-step edits, and alignment with complex textual instructions.
(2) \textbf{ImgEdit} \citep{ye2025imgedit}, which assesses editing quality and consistency preservation by requiring models to modify specific regions or attributes while maintaining the overall structure, identity, and visual coherence of the original image.

Across these benchmarks, \method supports both single-stage evaluation (direct inference and scoring) and multi-stage pipelines (post-training and then evaluation), enabling flexible and realistic assessment protocols.

\section{Detailed Results}
\label{sec:detailed_results}

We report fine-grained subscore breakdowns across all evaluated benchmarks to provide a more detailed understanding of model performance under the unified evaluation framework implemented in TorchUMM.

Specifically, we present per-dimension subscores for generation, understanding and editing tasks across diverse benchmarks, including GenEval \Cref{tab:sub-geneval}, WISE \Cref{tab:sub-wise}, MathVista \Cref{tab:sub-mathvista}, MMMU \Cref{tab:sub-mmmu}, GEdit-EN \Cref{tab:sub-gedit-en-in,tab:sub-gedit-en-o}, RealUnify UEG score \Cref{tab:sub-realunify} uni-MMMU \Cref{tab:uni-mmmu}.

\input{tab/app-sub-geneval}
\input{tab/app-sub-wise}

\input{tab/app-sub-mathvista}
\input{tab/app-sub-mmmu}

\input{tab/app-sub-gedit-en-o}
\input{tab/app-sub-gedit-en-in}
\input{tab/app-sub-realunify}
\input{tab/app-uni-mmmu}

\section{Detailed Unification Analysis}
\label{sec-append-analysis}

\subsection{Unification Degree is Not a Reliable Capability Proxy}
\label{sec-app-analysis-ueval}

To examine whether tighter architectural unification actually leads to stronger behavior on tightly coupled understanding--generation tasks, we present a focused UEval~\citep{li2026ueval} case study. We use \textit{degree of unification} as a coarse architectural notion, referring to how much multimodal representation, generation pathway, and model parameters are shared across modalities and tasks. Under this view, MMaDA is the most unified of the three: it models different modalities as a single token sequence and performs generation within a diffusion language model. Show-o2 is intermediate: it retains a unified token space, but routes text and image generation through different heads. OmniGen2 is the least unified: it follows a modular orchestration design in which a VLM supplies condition tokens to a separate visual generator. This yields a natural prior ordering in unification degree, namely MMaDA $>$ Show-o2 $>$ OmniGen2.

\begin{figure}[!t]
\centering
\parbox[t]{0.98\textwidth}{\small\textbf{Sample 4 (art).} \textit{How to draw a cartoon dog? Show each step both visually and textually.}}\\[0.3em]
\includegraphics[width=0.94\textwidth]{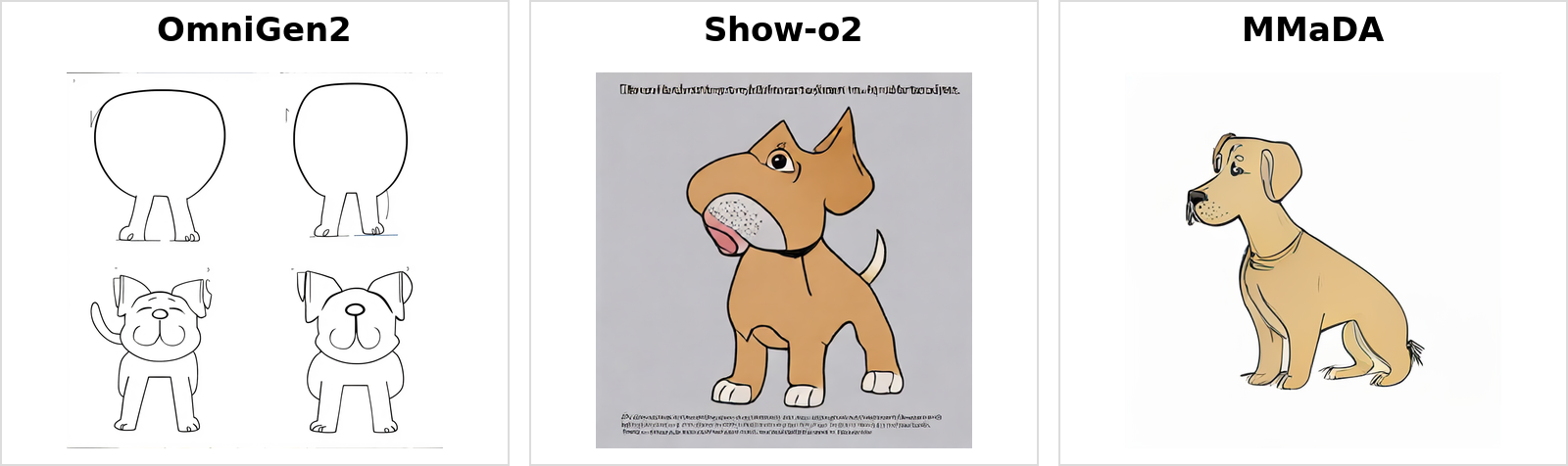}\\[0.6em]
\parbox[t]{0.98\textwidth}{\small\textbf{Sample 481 (paper).} \textit{Draw the overall Transformer encoder--decoder structure and explain how data flows through the layers.}}\\[0.3em]
\includegraphics[width=0.94\textwidth]{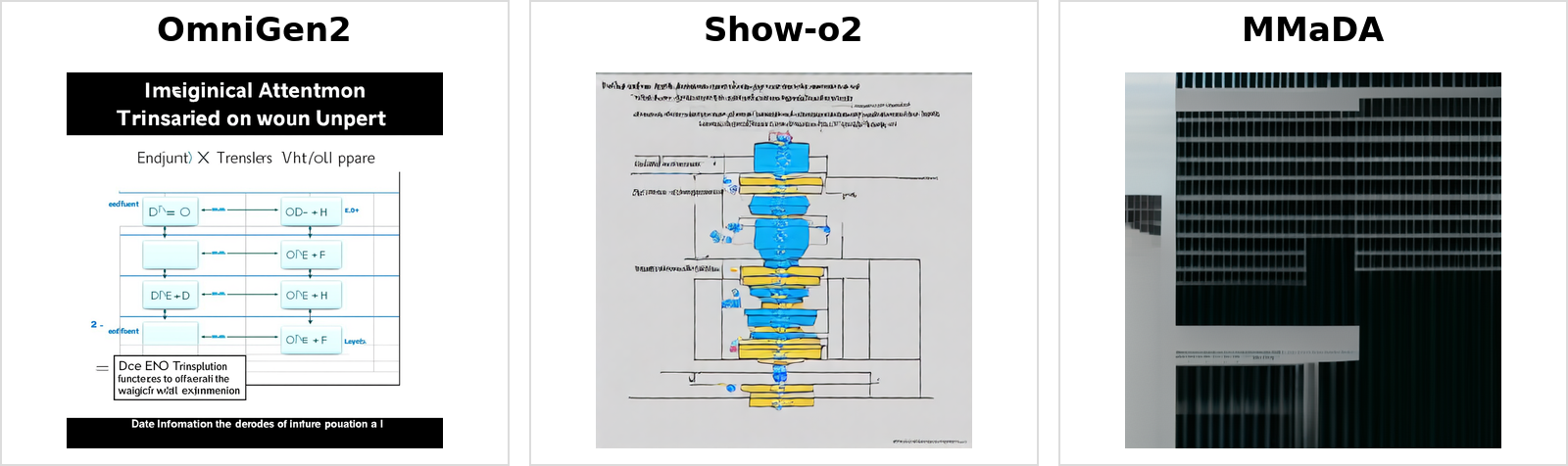}
\caption{Representative UEval cases across models with different degrees of unification. The first row requires procedural decomposition, where the model must align textual guidance with a progressive visual sequence. The second row requires technical understanding, structured layout, arrows, and text rendering in a scientific diagram. Across both cases, only OmniGen2 consistently attempts the requested structure, while Show-o2 tends to produce more visually plausible but semantically under-specified outputs and MMaDA more often collapses on both semantic grounding and visual organization.}
\label{fig:app-ueval-case-study}
\end{figure}
\FloatBarrier

\Cref{fig:app-ueval-case-study} highlights two representative prompts that probe different kinds of multimodal coupling. Sample~4 from the \textit{art} subset asks for a beginner-friendly, step-by-step drawing tutorial. To succeed, a model must decompose the task into an ordered sequence, align each intermediate visual state with its textual explanation, and preserve cumulative progression across steps. Sample~481 from the \textit{paper} subset tests a different capability profile: abstract technical understanding of the Transformer, explicit encoder--decoder layout, data-flow arrows, and legible textual labels. Yet the relative ordering is the same in both cases. Only OmniGen2 makes a consistent attempt to satisfy the prompt at the intended level of precision. On Sample~4, it produces multiple progressive panels and achieves a much higher rubric-based score than Show-o2 and MMaDA (0.79 vs.\ 0.46 and 0.29). On Sample~481, all three models struggle, but OmniGen2 still produces the only output that resembles a structured diagram of the requested architecture (0.13 vs.\ 0.07 and 0.00).

Taken together, these cases suggest that the degree of unification is not a reliable proxy for effective model capability. In our comparison, the observed capability ordering is nearly the reverse of the architectural prior: OmniGen2 outperforms Show-o2 and MMaDA on both cross-modal understanding and final visual presentation, despite being the least unified system among the three. At the same time, this should not be interpreted as a clean causal estimate of unification itself. The models also differ substantially in backbone inheritance and generative formulation. MMaDA builds on LLaDA-8B-Instruct and adopts a block-masked diffusion language modeling objective. Show-o2 inherits from Qwen2.5-7B-Instruct and combines a shared backbone with separate LM and flow heads. OmniGen2 uses a Qwen2.5-3B-Instruct VLM to condition a separate visual generator. Their training data composition and quality are likewise uncontrolled. The point, therefore, is not that unified modeling is unhelpful. Rather, the practical benefits of unification remain far from fully unlocked, and are currently entangled with, and often dominated by, differences in architecture, inherited pretraining, optimization recipe, and data quality. One plausible explanation is that tighter unification also introduces harder cross-modal and cross-task interference, so without sufficiently strong representations and data, the optimization burden can outweigh the theoretical benefits of a shared model. This motivates future work that varies the unification mechanism under strict controls over backbone architecture, initialization, tokenizer, and data recipe. Such controlled comparisons are essential for identifying what unification itself changes in model behavior, rather than conflating it with broader differences in model design and pretraining.

\subsection{Heterogeneous Effects of Unification Training}
\label{sec-app-analysis-query-consistency}

The UEval case study above already suggests that more architectural unification does not automatically translate into better downstream behavior. We now ask a sharper question: when a unified model is initialized from a strong \emph{non-unified} backbone and then jointly optimized across multimodal tasks, how much of the backbone's original behavior is preserved, and how much is rewritten by unification training? We use robustness to semantically equivalent query variations as a diagnostic lens. If unification training only lightly perturbs the inherited backbone, behavior under small meaning-preserving prompt changes should remain close to that of the initializer; if it reshapes the backbone more aggressively, the drift should become visible in both outputs and latent states.

Starting from the same 200-sample MathVista subset, we use Qwen3.5-9B~\citep{qwen2026qwen35} in non-thinking mode to generate two meaning-preserving rephrasings for each original question, yielding three text variants per image. We then run deterministic inference for each model on all three variants, embed both the queries and the resulting responses with Qwen3-VL-Embedding-8B~\citep{qwen2026qwen3vlembedding}, and compare them in a shared semantic space via pairwise \emph{cosine similarity}. To probe the models' own internal behavior, we re-feed the complete multimodal trace, extract hidden states every five layers, and summarize each layer by the mean representation over the response span. This gives us two aligned views of the effect of unification training: external response embeddings and internal latent dynamics.

\begin{figure}[!htbp]
\centering
\setlength{\tabcolsep}{2pt}
\begin{tabular}{cccc}
\multicolumn{2}{c}{\textbf{OmniGen2}} & \multicolumn{2}{c}{\textbf{Qwen2.5-VL-3B-Instruct}} \\
\multicolumn{1}{c}{\footnotesize External response embeddings} & \multicolumn{1}{c}{\footnotesize Internal latent dynamics} &
\multicolumn{1}{c}{\footnotesize External response embeddings} & \multicolumn{1}{c}{\footnotesize Internal latent dynamics} \\
\includegraphics[width=0.24\textwidth]{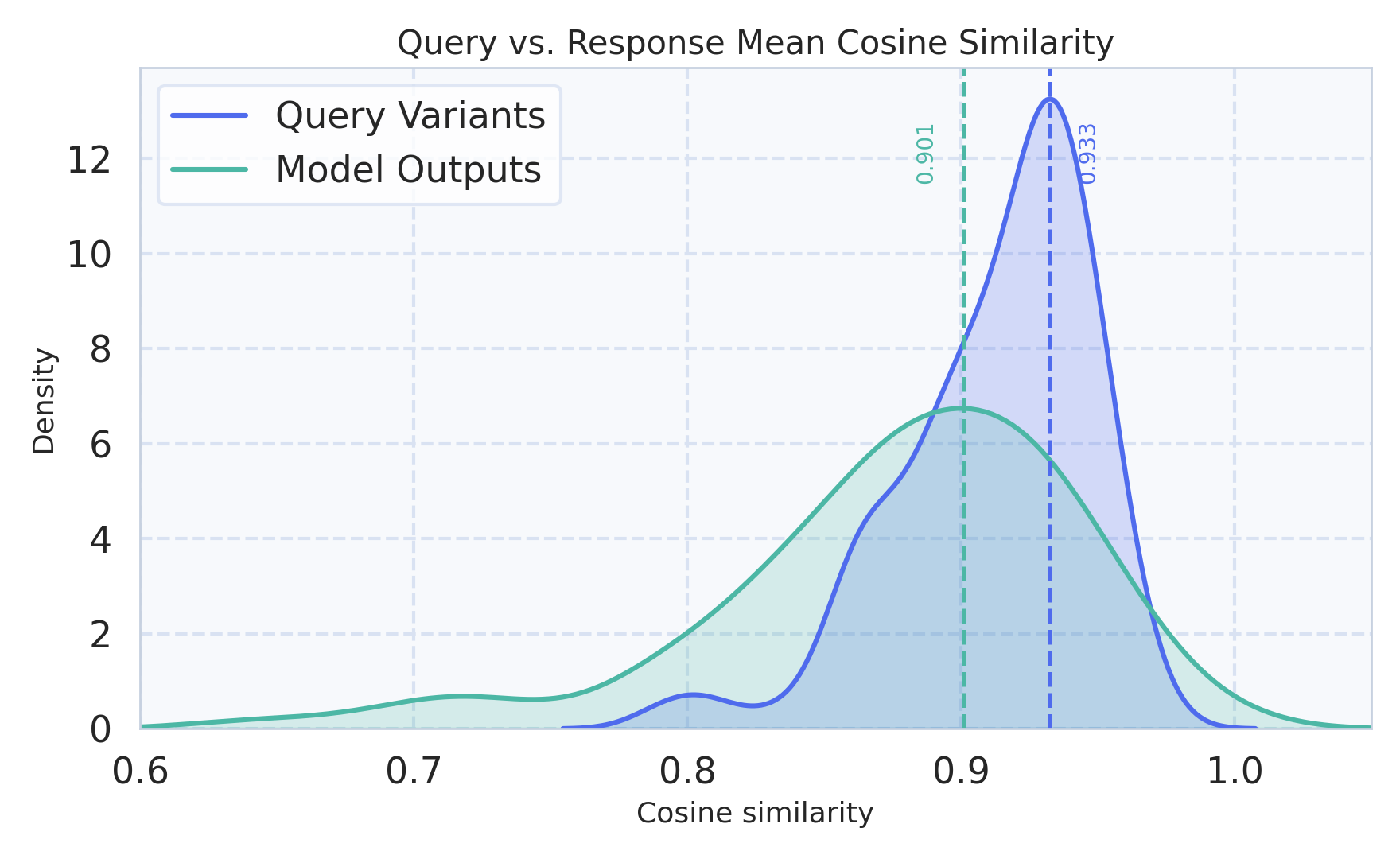} &
\includegraphics[width=0.24\textwidth]{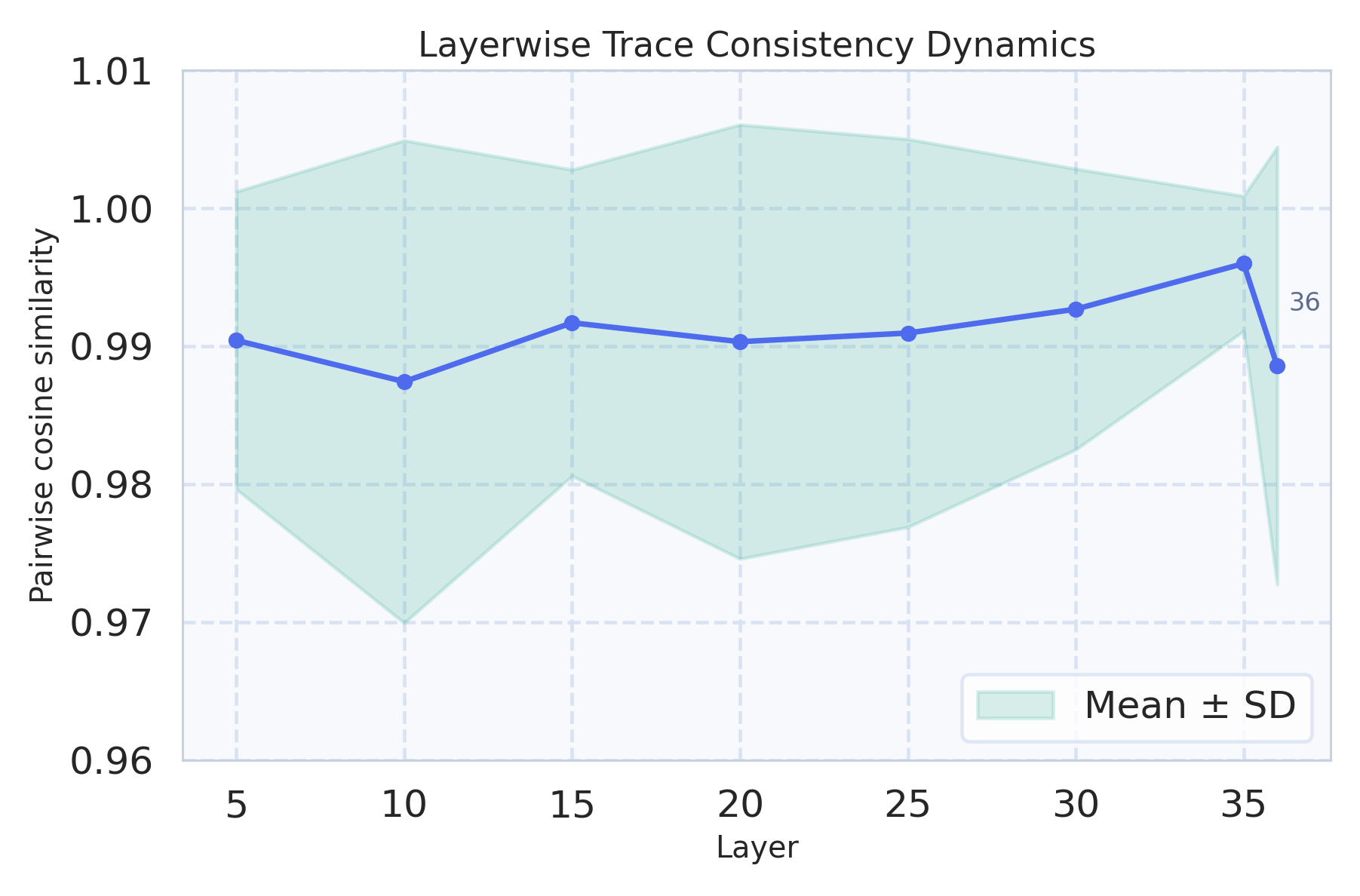} &
\includegraphics[width=0.24\textwidth]{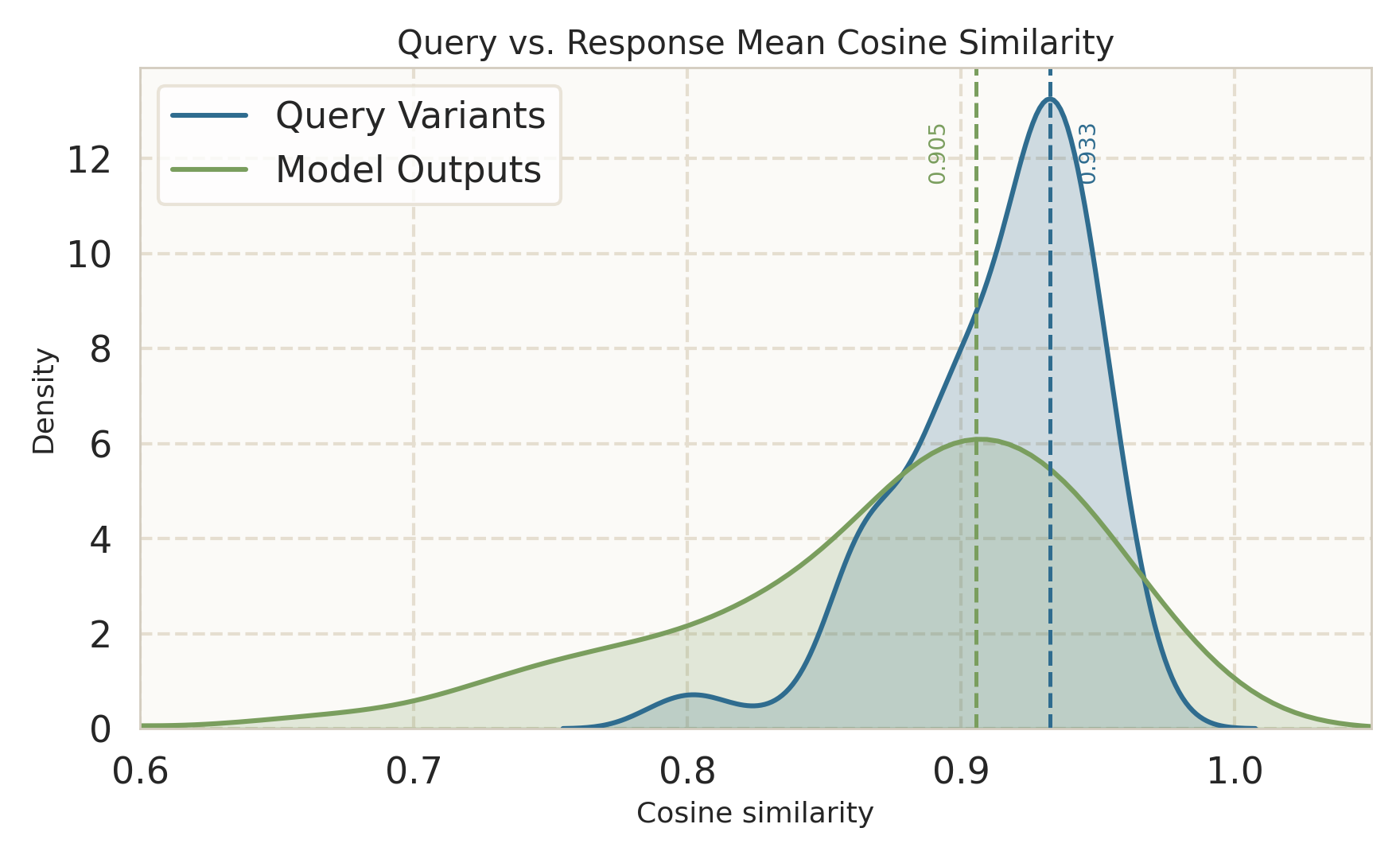} &
\includegraphics[width=0.24\textwidth]{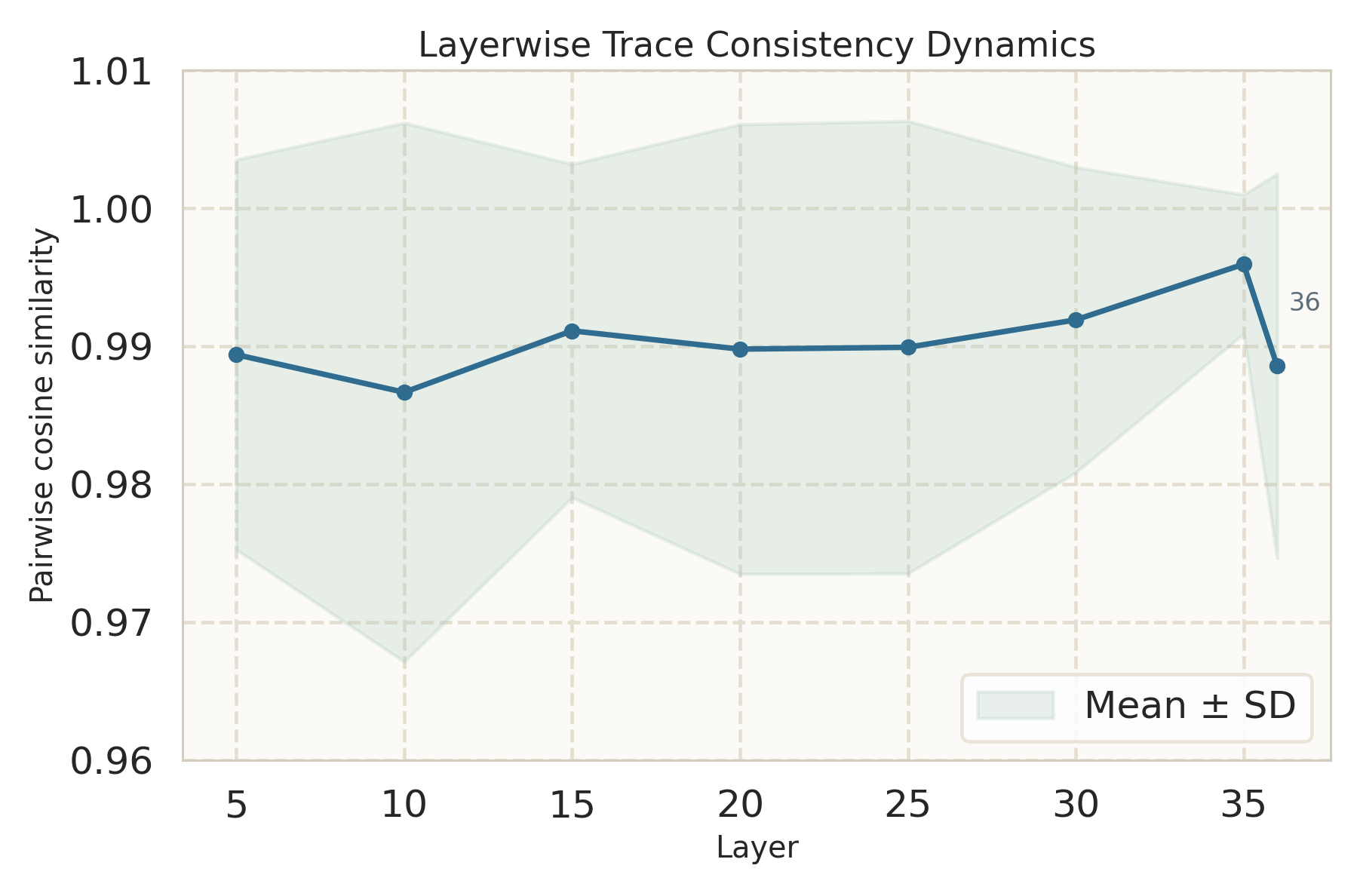} \\
\\[-0.2em]
\multicolumn{2}{c}{\textbf{Show-o2}} & \multicolumn{2}{c}{\textbf{Qwen2.5-VL-7B-Instruct}} \\
\multicolumn{1}{c}{\footnotesize External response embeddings} & \multicolumn{1}{c}{\footnotesize Internal latent dynamics} &
\multicolumn{1}{c}{\footnotesize External response embeddings} & \multicolumn{1}{c}{\footnotesize Internal latent dynamics} \\
\includegraphics[width=0.24\textwidth]{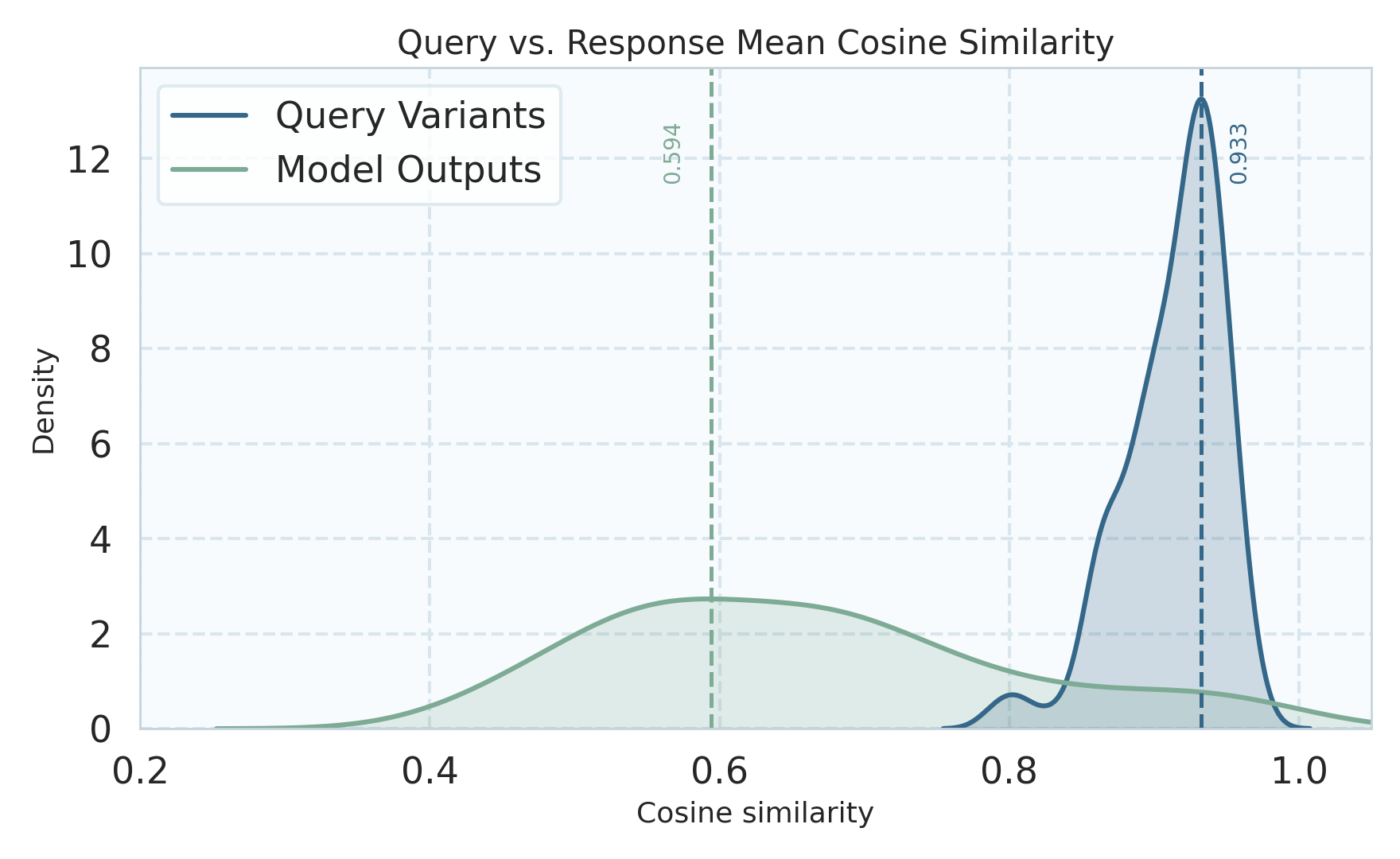} &
\includegraphics[width=0.24\textwidth]{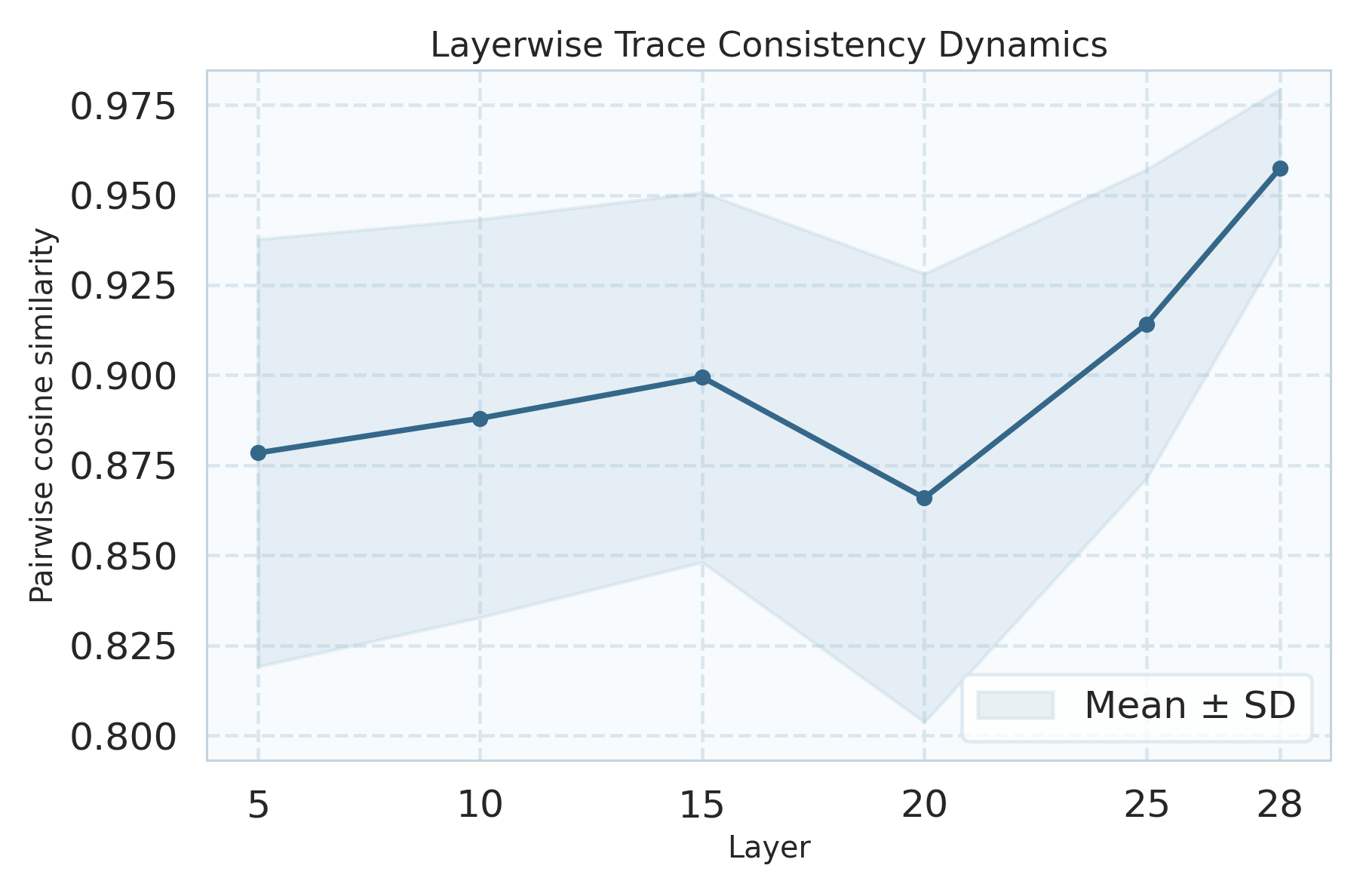} &
\includegraphics[width=0.24\textwidth]{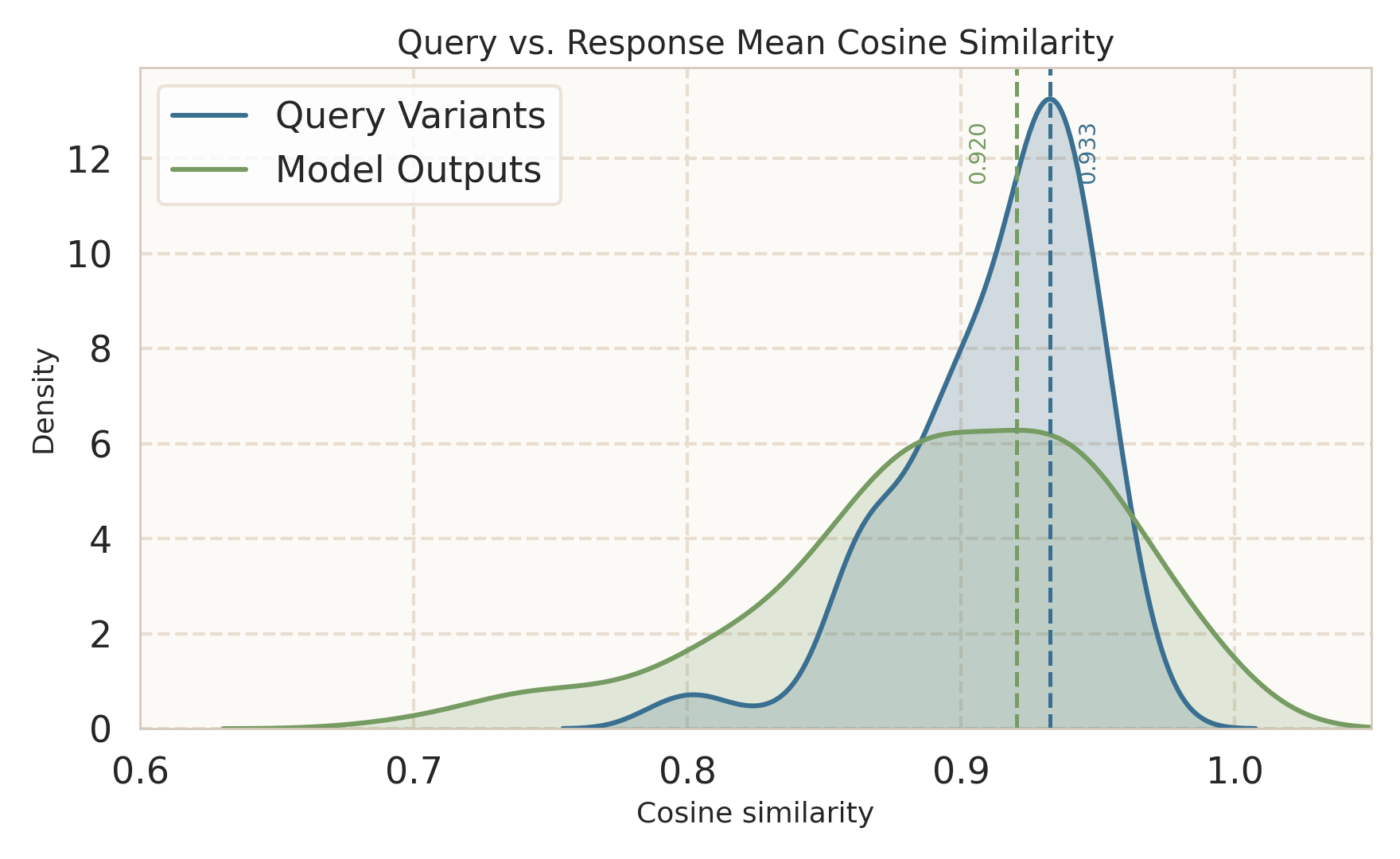} &
\includegraphics[width=0.24\textwidth]{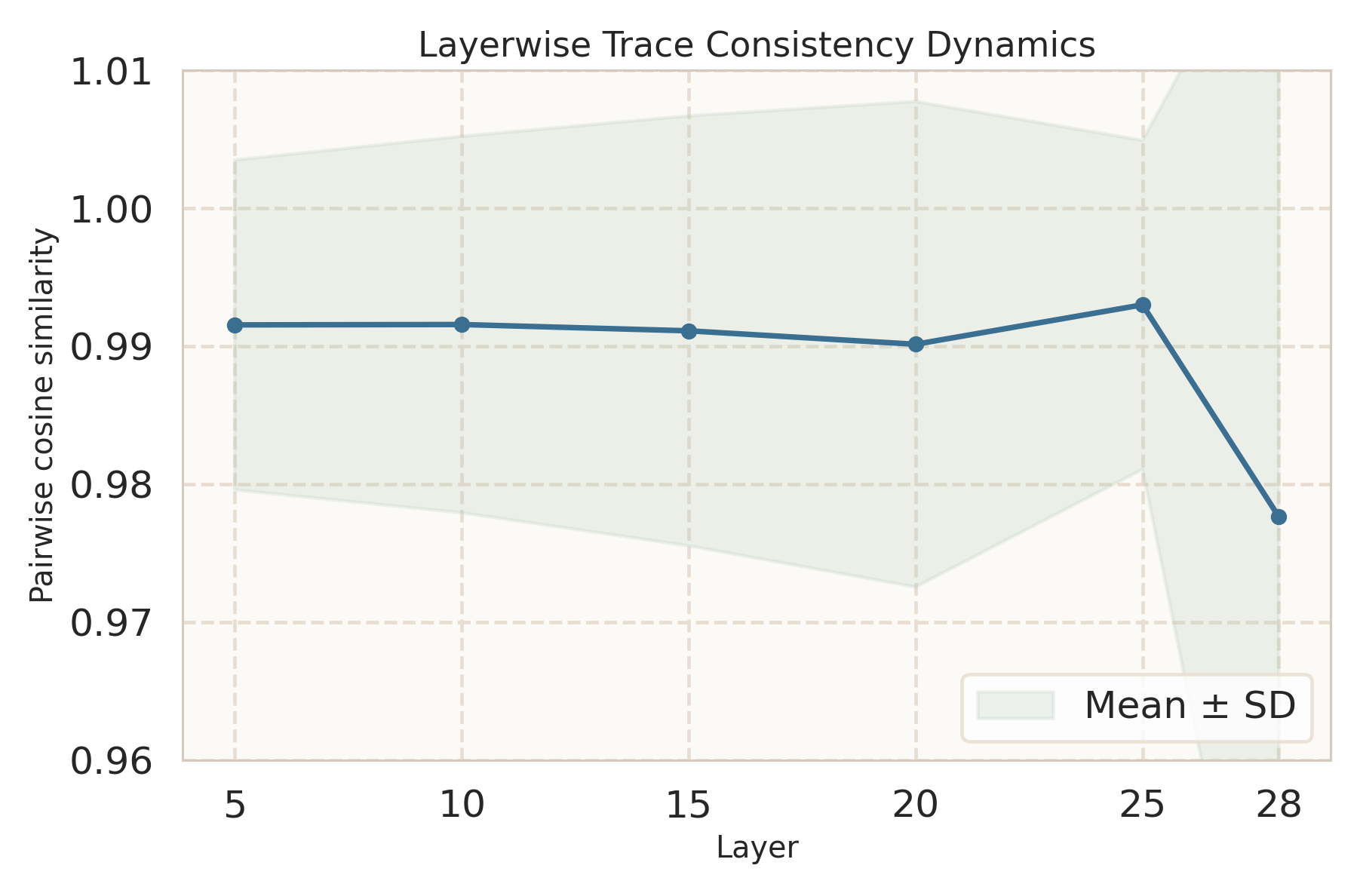}
\end{tabular}
\caption{Query-variation analysis under two backbone--model pairings. In each row, the left panel shows consistency among the generated responses in a shared embedding space, while the right panel shows consistency of the corresponding response representations across layers. The first row compares the relatively low-unification pair OmniGen2 and Qwen2.5-VL-3B-Instruct; the second row compares the more aggressively unified Show-o2 and its Qwen2.5-VL-7B-Instruct reference. Read row-wise, the figure asks how strongly unification training has altered the inherited backbone behavior.}
\label{fig:app-analysis-query-consistency}
\end{figure}
\FloatBarrier

\Cref{fig:app-analysis-query-consistency} makes the contrast between the two unification regimes visually explicit. In the lower-unification pair, OmniGen2 remains behaviorally very close to Qwen2.5-VL-3B-Instruct, the backbone from which it is initialized. Their response-consistency curves in embedding space nearly overlap, and their layerwise latent trajectories are almost indistinguishable: both stay near the top of the scale throughout the network, with only tiny fluctuations. Through this lens, OmniGen2's unification recipe looks largely behavior-preserving. It becomes a unified multimodal model, but it does not seem to rewrite the inherited backbone's response to small meaning-preserving perturbations very much.

The higher-unification pair behaves differently. Qwen2.5-VL-7B-Instruct retains the high-stability profile we would expect from a strong pretrained backbone: the response-consistency density is tight and concentrated, and the latent trajectory remains almost flat at a very high level across layers. Show-o2, by contrast, exhibits a much broader and flatter output distribution, with substantial mass spread into the low-similarity region. The same qualitative change appears internally: rather than tracking the nearly flat latent curve of the Qwen backbone, Show-o2 starts markedly lower, remains less consistent through the middle layers, and only partially recovers near the end. This suggests that the higher-degree unification used by Show-o2 has changed the inherited backbone behavior substantially, not merely reformatted its outputs.

Taken together, the two comparisons refine the message of \Cref{sec-app-analysis-ueval}. The key issue is not simply whether a model is ``more unified,'' but what that unification training does to the backbone it starts from. Under a relatively modular recipe, unification can leave the initializer's behavior largely intact. Under a more strongly shared recipe, it can reshape that behavior much more aggressively. Crucially, stronger change is not automatically better: in our setting, the more heavily unified model is also the one whose consistency degrades more severely relative to its backbone. A natural interpretation is that tighter sharing across understanding and generation introduces stronger cross-task interference, so the effect of unification depends not only on architectural elegance, but also on whether the optimization recipe and data mixture are good enough to keep that interference under control. Consistency is useful here precisely because it exposes this hidden side of unification training: not just what capability is gained, but what inherited behavior is preserved, altered, or destabilized along the way.

\subsection{Reasoning Traces Expose a Test-Time Scaling Gap}
\label{sec-app-analysis-reasoning}

We next examine a complementary behavior: whether UMMs can follow an explicit chain-of-thought (CoT) instruction and convert additional visible test-time computation into improved answer quality. MathVista questions often require reading the image, selecting relevant visual evidence, performing intermediate symbolic or spatial reasoning, and only then committing to an answer. A model that has preserved the reasoning behavior of its language or VLM backbone should therefore be able to externalize a structured reasoning trace when prompted to do so. Conversely, if unification training disrupts this capability, the model may collapse to short answer-style responses even under CoT prompting.

We evaluate MMaDA, Show-o2, and OmniGen2 on the same 200-sample MathVista subset used in \Cref{sec-app-analysis-query-consistency}, using the original questions and deterministic CoT prompting. For reference, we also include GPT-5.4 on the same samples, but only as an answer-quality reference: the API does not expose hidden reasoning tokens, so its visible response length should not be interpreted as a complete CoT budget. For each trace, we measure answer accuracy, visible token length, and teacher-forced next-token entropy along the generated response for the open-source models.

\FloatBarrier

\Cref{fig:app-analysis-reasoning} reveals a sharp separation among the three UMMs. OmniGen2 produces explicit CoT traces for almost all samples (99.5\%), with a median visible length of 220 tokens and the highest open-source accuracy (58.1\%). Its entropy trajectory also exhibits a relatively healthy pattern: uncertainty starts at a moderate level and decreases gradually over the trace, suggesting that the model is progressively constraining the answer as reasoning unfolds. In contrast, MMaDA and Show-o2 rarely make effective use of the CoT instruction. MMaDA has a median trace length of only 2 tokens and emits answer-only responses on 63.1\% of samples; Show-o2 is slightly less extreme but still has a median trace length of only 5 tokens and emits answer-only responses on 49.5\% of samples. Their accuracies, 29.3\% and 50.0\%, lag behind OmniGen2 despite their higher architectural unification.

Thus, the failure mode is not merely lower final accuracy; it is a trace-formation gap. The more aggressively unified models often fail to preserve or activate the backbone's CoT-following behavior, so additional test-time reasoning instructions do not translate into the kind of visible reasoning process that MathVista rewards.

This result strengthens the message of \Cref{sec-app-analysis-ueval,sec-app-analysis-query-consistency}. In \Cref{sec-app-analysis-ueval}, the more unified models struggled to decompose and control tightly coupled text-to-image instructions. In \Cref{sec-app-analysis-query-consistency}, stronger unification training could destabilize inherited consistency. Here, the same pattern appears in reasoning: stronger sharing does not automatically yield better integration of perception, language, and intermediate computation. Instead, without careful preservation of the underlying reasoning capability, unification can leave the model unable to exploit CoT-style test-time scaling.

\subsection{Unification Does Not Necessarily Reduce the Modality Gap}
\label{sec-app-analysis-latent-gap}

The preceding subsections expose three behavioral symptoms: weaker text-to-image instruction control, degraded consistency relative to the inherited backbone, and limited CoT-following ability. We now ask whether these symptoms have a common representational signature. If architectural unification truly produced a more unified multimodal model, we would expect paired text and image representations to become closer in the model's own latent space. Conversely, if aggressive cross-modal and cross-task sharing creates interference before the model has sufficient scale, data, or optimization support, then a model that is more unified by design may still exhibit a larger \emph{modality gap} internally.

We study this question on UEval, where each sample naturally couples a text-side interpretation of a T2I instruction with the corresponding generated image. For each model, we re-encode its own \((\text{prompt}, \text{image answer}, \text{text answer})\) triplet through the multimodal understanding pathway, extract pooled hidden representations over the text-answer span and image-answer span at selected layers, and visualize the final-layer representations. This setup is useful because the paired spans reflect both cross-modality alignment (text versus image) and cross-task continuity (understanding the instruction before producing the image).

To make this comparison quantitative, we introduce the \emph{Cross-Modal Alignment Margin} (CMAM). For a fixed layer $\ell$, let $\hat{z}^{\ell}_{t,i}$ be the L2-normalized pooled representation of sample $i$'s text-answer span, and let $\hat{z}^{\ell}_{v,i}$ be the corresponding normalized image-answer representation. We compute
\begin{equation}
\mathrm{CMAM}^{\ell}
= \frac{1}{N}\sum_{i=1}^{N}
\cos\!\left(\hat{z}^{\ell}_{t,i}, \hat{z}^{\ell}_{v,i}\right)
- \frac{1}{N(N-1)}\sum_{i\neq j}
\cos\!\left(\hat{z}^{\ell}_{t,i}, \hat{z}^{\ell}_{v,j}\right).
\end{equation}
CMAM measures how much closer a sample's own text--image pair is than mismatched text--image pairs from other samples. A higher CMAM indicates that paired text and image outputs occupy a more coherent shared latent space; a lower CMAM indicates that the paired modalities are not much more aligned than random mismatches, suggesting a larger modality gap.

\begin{figure}[!htbp]
\centering
\setlength{\tabcolsep}{3pt}
\begin{tabular}{ccc}
\textbf{MMaDA-8B} & \textbf{Show-o2-7B} & \textbf{OmniGen2-8B} \\
\includegraphics[width=0.315\textwidth]{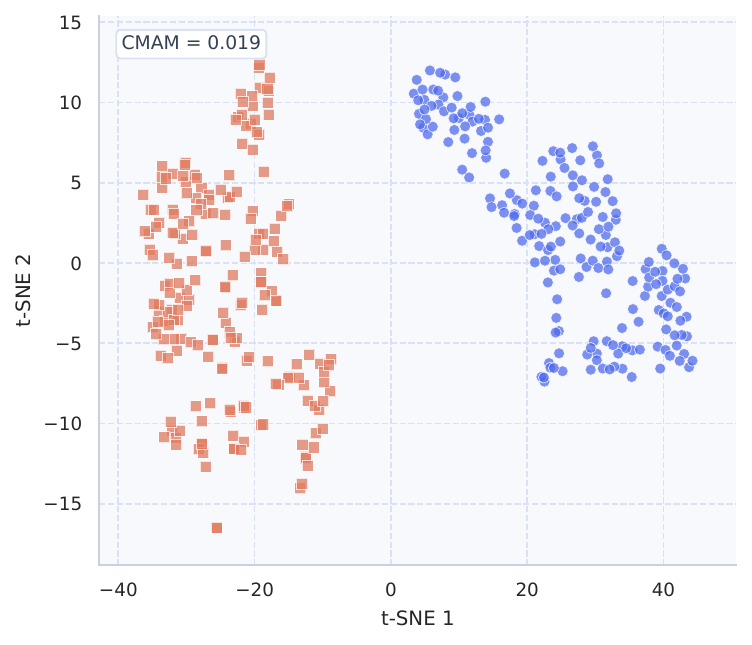} &
\includegraphics[width=0.315\textwidth]{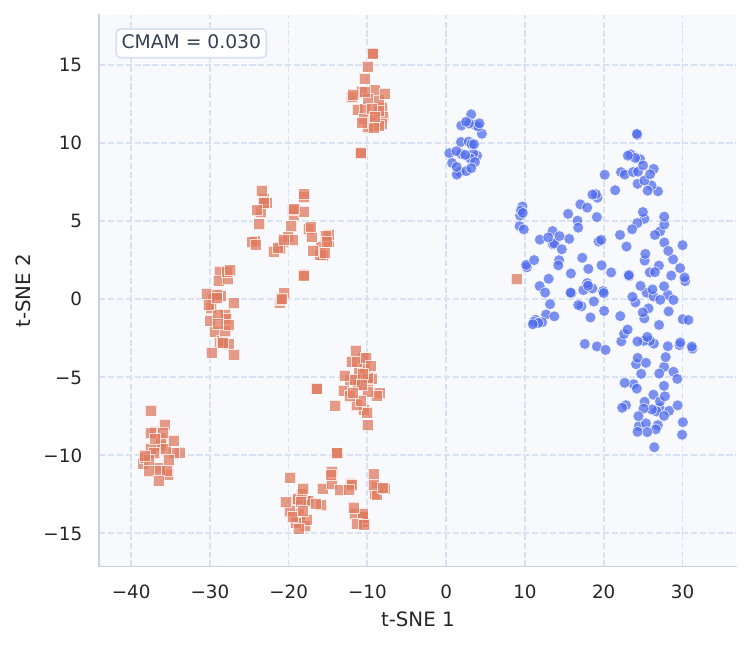} &
\includegraphics[width=0.315\textwidth]{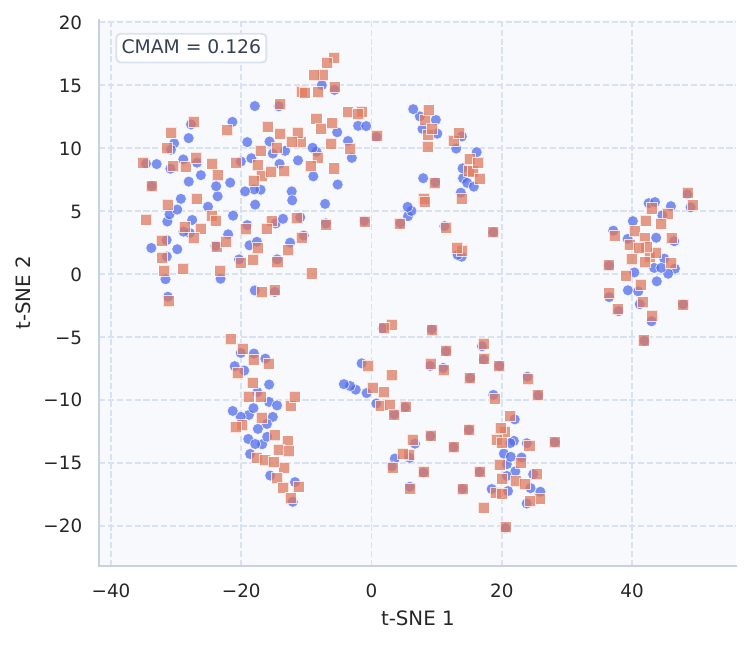}
\end{tabular}\\[-0.3em]
\includegraphics[width=0.28\textwidth]{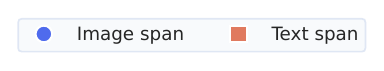}
\caption{Final-layer UEval latent spaces under different unification regimes. Each point is a pooled representation from the model's own multimodal trace; blue circles denote image spans and orange squares denote text spans. We report Cross-Modal Alignment Margin (CMAM) in each panel. Despite being the least unified by architectural design, OmniGen2 exhibits the strongest text--image alignment and the smallest effective modality gap. MMaDA and Show-o2, which impose more aggressive cross-modal and cross-task sharing, show substantially weaker paired alignment.}
\label{fig:app-analysis-latent-gap}
\end{figure}
\FloatBarrier

\Cref{fig:app-analysis-latent-gap} shows that the learned representation space does not follow the architectural unification prior. OmniGen2, the least unified system by design, has the largest final-layer CMAM (0.126) and visually interleaves paired text and image spans throughout the visualized manifold. Show-o2 and MMaDA, despite being more unified architecturally, have much smaller final-layer CMAM values (0.030 and 0.019). Their paired text and image representations are therefore less distinguishable from mismatched pairs, indicating weaker cross-modal alignment in the model's own latent space.

This provides a unifying explanation for the earlier behavioral results. OmniGen2 preserves a smaller effective modality gap, which is consistent with its stronger UEval instruction following in \Cref{sec-app-analysis-ueval}, its more stable backbone-like behavior in \Cref{sec-app-analysis-query-consistency}, and its better CoT-following behavior in \Cref{sec-app-analysis-reasoning}. By contrast, MMaDA and Show-o2 embody stronger architectural sharing, but that sharing does not yet translate into a more aligned representation space. Instead, it may intensify cross-modal and cross-task interference: the same parameters must support visual generation, language reasoning, multimodal understanding, and trace formation before the learned latent geometry is sufficiently organized to make those capabilities reinforce one another.

The broader conclusion is therefore not that unification is undesirable, but that \emph{architectural unification is not equivalent to representational unification}. Before some scale, data, or optimization threshold is reached, more aggressive parameter sharing can increase the practical modality gap rather than reduce it. Understanding how to make unification actually produce a smaller modality gap and stronger downstream capability will require controlled studies that fix the backbone, tokenizer, data mixture, and optimization recipe while varying only the unification mechanism. TorchUMM is intended to make precisely this kind of controlled analysis possible.

\section{AI Assistants Usage}
\label{sec-app-llm}
AI assistants were used as auxiliary tools in preparing this manuscript, primarily for language refinement, clarity and organization. The experimental design and methodological choices were made by the authors.

\section{Broader Impact}
\label{sec-appendix-impact}

This work presents \method, a unified codebase for evaluating and analyzing unified multimodal models (UMMs). By standardizing evaluation pipelines across models and tasks, it may facilitate more consistent and reproducible research in multimodal learning.

The framework could support the development of more capable multimodal systems by enabling systematic comparison and analysis. At the same time, like other tools for improving model performance, it may indirectly contribute to stronger generative models, which could be misused for producing misleading or low-quality content at scale.

In addition, the framework relies on existing datasets and models, and therefore may inherit their limitations, including potential biases or incomplete coverage of real-world scenarios. Evaluation results should thus be interpreted with caution and complemented with additional analysis when applied in practice.

Overall, \method is intended as a research-oriented toolkit, and its broader impact depends on how it is used by the community.


%% file: tab/app-sub-geneval.tex
\begin{table}[H]
\centering
\caption{Geneval sub-score. 
}
\begin{tabular}{l|ccccccc}
model & \multicolumn{1}{l}{single\_object} & \multicolumn{1}{l}{two\_object} & \multicolumn{1}{l}{counting} & \multicolumn{1}{l}{colors} & \multicolumn{1}{l}{position} & \multicolumn{1}{l}{color\_attr} & \multicolumn{1}{l}{overall} \\ \midrule
bagel(w/o think) & 99.38 & 94.19 & 78.75 & 87.77 & 51 & 61.75 & 78.81 \\
blip3o & 98.12 & 93.18 & 73.44 & 86.17 & 72.75 & 64.5 & 81.36 \\
show\_o2(7B) & 97.81 & 71.46 & 48.75 & 78.46 & 20 & 42.75 & 59.87 \\
show\_o2(1.5B) & 96.88 & 64.39 & 46.88 & 76.06 & 16.75 & 32 & 55.49 \\
Janus\_pro & 97.81 & 86.62 & 57.5 & 89.36 & 76 & 66.25 & 78.92 \\
Janus & 85.62 & 37.63 & 18.75 & 53.46 & 17.5 & 27.25 & 40.04 \\
janus\_flow & 94.25 & 46.06 & 27.75 & 74.68 & 32.2 & 25 & 49.99 \\
omnigen2 & 99.69 & 93.94 & 68.75 & 88.03 & 53.25 & 67.5 & 78.53 \\
tokenflow & 97.19 & 59.6 & 37.81 & 86.17 & 17.25 & 15.25 & 52.21 \\
Emu3 & 94.69 & 55.81 & 30 & 76.06 & 8.5 & 9.5 & 45.76 \\
deepgen & 98.75 & 98.99 & 81.25 & 92.55 & 75 & 73 & 86.59 \\
mmada & 89.06 & 49.75 & 31.25 & 73.67 & 12.5 & 20.5 & 46.12 \\
emu3.5 & 100 & 93.94 & 49.06 & 91.49 & 85.5 & 71 & 81.83 \\
ovis\_u1 & 100 & 94.95 & 93.75 & 93.62 & 80 & 78 & 90.05 \\
bagel\_IRG & 98.44 & 87.37 & 70.31 & 78.72 & 40.5 & 57 & 72.06 \\
bagel\_recA & 99.38 & 94.44 & 79.38 & 89.1 & 61.75 & 74.25 & 83.05 \\
bagel\_recA-ema & 99.06 & 95.71 & 78.12 & 84.84 & 53.25 & 62.25 & 78.87 \\
bagel\_unigame & 98.44	& 95.96	& 81.25	& 93.62	& 72.00 & 75.75 & 86.17 \\
bagel\_sft & 99.06 & 92.42 & 77.5 & 86.97 & 49.75 & 62.5 & 78.03 \\
blip3o\_sft & 99.06 & 91.16 & 69.38 & 84.84 & 69.25 & 56.75 & 78.41 \\
janus\_pro\_sft & 97.5 & 87.63 & 55.94 & 89.36 & 73 & 62.25 & 77.61 \\
omnigen2\_sft & 99.38 & 93.18 & 71.56 & 86.17 & 53.5 & 63.25 & 77.84 \\
show\_o2\_sft & 94.38 & 55.56 & 32.19 & 77.93 & 18.75 & 34 & 52.13 \\
tokenflow\_sft & 95.31 & 56.57 & 37.81 & 80.59 & 15.75 & 25.75 & 51.96
\end{tabular}
\label{tab:sub-geneval}
\end{table}

%% file: tab/app-sub-wise.tex
\begin{table}[t]
\centering
\caption{Wise generation sub-score. 
} 
\begin{tabular}{l|ccccccc}

\label{tab:app-sub-wise}
model & \multicolumn{1}{l}{culture} & \multicolumn{1}{l}{time} & \multicolumn{1}{l}{space} & \multicolumn{1}{l}{biology} & \multicolumn{1}{l}{physics} & \multicolumn{1}{l}{chemistry} & \multicolumn{1}{l}{overall} \\ \midrule
bagel(w/o think) & 0.3883 & 0.4386 & 0.4714 & 0.362 & 0.4205 & 0.294 & 0.3989 \\
blip3o & 0.4028 & 0.4186 & 0.5259 & 0.4025 & 0.4255 & 0.3 & 0.4138 \\
show\_o2(7B) & 0.3641 & 0.3497 & 0.4519 & 0.3455 & 0.369 & 0.239 & 0.3595 \\
show\_o2(1.5B) & 0.3111 & 0.3563 & 0.4357 & 0.315 & 0.384 & 0.231 & 0.3349 \\
show\_o & 0.2865 & 0.3225 & 0.4132 & 0.275 & 0.33 & 0.198 & 0.3037 \\
Janus\_pro & 0.3616 & 0.3853 & 0.4789 & 0.3605 & 0.4745 & 0.2485 & 0.3811 \\
janus & 0.208 & 0.2707 & 0.3508 & 0.1705 & 0.191 & 0.1095 & 0.2222 \\
janus\_flow & 0.2731 & 0.3222 & 0.3947 & 0.3215 & 0.286 & 0.1905 & 0.2964 \\
omnigen2 & 0.418 & 0.4042 & 0.4887 & 0.3635 & 0.3875 & 0.281 & 0.4029 \\
tokenflow & 0.3253 & 0.3626 & 0.3357 & 0.2915 & 0.2605 & 0.151 & 0.3056 \\
emu3 & 0.3463 & 0.3482 & 0.3711 & 0.331 & 0.3685 & 0.213 & 0.3373 \\
deepgen & 0.5989 & 0.4955 & 0.6102 & 0.4765 & 0.5515 & 0.408 & 0.547 \\
emu3.5 & 0.7001 & 0.5683 & 0.6944 & 0.6435 & 0.6085 & 0.406 & 0.6331 \\
ovis\_u1 & 0.3643 & 0.3991 & 0.4831 & 0.3405 & 0.409 & 0.239 & 0.3755 \\
MMaDA & 0.6502 & 0.6814 & 0.7492 & 0.662 & 0.742 & 0.4205 & 0.656 \\
bagel\_recA & 0.4035 & 0.4147 & 0.5432 & 0.3985 & 0.463 & 0.334 & 0.4225 \\
bagel\_IRG & 0.3674 & 0.4081 & 0.465 & 0.3575 & 0.4495 & 0.2655 & 0.3842 \\
bagel\_sft & 0.2201 & 0.2117 & 0.3466 & 0.173 & 0.2545 & 0.151 & 0.2274 \\
bagel\_unicot & 0.3998 & 0.4183 & 0.4797 & 0.3405 & 0.455 & 0.306 & 0.4037 \\
januspro\_unigame & 0.34 & 0.3769 & 0.4789 & 0.374 & 0.4722 & 0.248 & 0.3729 \\
bagel\_unigame & 0.3956 & 0.4138 & 0.4876 & 0.359 & 0.4355 & 0.3155 & 0.4032 \\
blip3o\_sft & 0.3926 & 0.4042 & 0.5053 & 0.3785 & 0.4175 & 0.274 & 0.3988 \\
show\_o2\_sft & 0.3121 & 0.3162 & 0.3966 & 0.315 & 0.3635 & 0.2345 & 0.3217 \\
omnigen2\_sft & 0.423 & 0.4117 & 0.4823 & 0.3835 & 0.388 & 0.2605 & 0.4053 \\
januspro\_sft & 0.35 & 0.3635 & 0.4526 & 0.399 & 0.467 & 0.228 & 0.3703 \\
tokenflow\_sft & 0.326 & 0.3539 & 0.45 & 0.262 & 0.3545 & 0.172 & 0.3282
\end{tabular}
\label{tab:sub-wise}
\end{table}

%% file: tab/app-sub-mathvista.tex
\begin{table}[t]
\centering
\caption{Mathvista sub-score. 
}
\begin{tabular}{l|ccc}
Model & \multicolumn{1}{l}{Overall} & \multicolumn{1}{l}{Multi-choice} & \multicolumn{1}{l}{Free-form} \\ \midrule
Bagel & 71.60\% & 80.19\% & 61.52\% \\
Show-o2 (7B) & 51.50\% & 63.52\% & 37.39\% \\
Show-o2 (1.5B) & 37.90\% & 53.33\% & 19.78\% \\
Show-o & 29\% & 44.07\% & 11.30\% \\
Emu3 & 44.90\% & 57.59\% & 30.00\% \\
Janus-Pro & 42.80\% & 51.30\% & 32.83\% \\
Emu3.5 & 30.60\% & 41.67\% & 17.61\% \\
MMaDA & 24.90\% & 38.70\% & 8.70\% \\
Ovis-U1 & 68.50\% & 75.74\% & 60.00\% \\
omnigen2\_sft & 63.5 & 71.67 & 53.91 \\
janus\_pro\_sft & 44.2 & 53.15 & 33.7 \\
bagel\_recA & 72.80\% & 79.81\% & 64.57\% \\
bagel\_unicot & 73 & 80.56 & 64.13 \\
bagel\_IRG & 68 & 73.52 & 61.52 \\
bagel\_unigame & 72.7 & 79.63 & 63.48 \\
bagel\_sft & 73.1 & 80 & 65
\end{tabular}
\label{tab:sub-mathvista}
\end{table}

%% file: tab/app-sub-mmmu.tex
\begin{table}[t]
\centering
\caption{MMMU sub-score. 
}
\resizebox{\linewidth}{!}{
\begin{tabular}{l|ccccccc}
Model & \multicolumn{1}{l}{Art \& Design} & \multicolumn{1}{l}{Business} & \multicolumn{1}{l}{Science} & \multicolumn{1}{l}{Health \& Medicine} & \multicolumn{1}{l}{Humanities \& Social Sci} & \multicolumn{1}{l}{Tech \& Engineering} & \multicolumn{1}{l}{Overall} \\ \midrule
Bagel & 0.583 & 0.433 & 0.433 & 0.587 & 0.725 & 0.438 & 0.519 \\
Janus-Pro & 0.508 & 0.38 & 0.347 & 0.413 & 0.5 & 0.352 & 0.407 \\
Janus & 0.325 & 0.18 & 0.267 & 0.273 & 0.3 & 0.3 & 0.273 \\
Janus-Flow & 0.392 & 0.233 & 0.233 & 0.333 & 0.358 & 0.243 & 0.29 \\
Emu3 & 0.342 & 0.26 & 0.267 & 0.34 & 0.392 & 0.31 & 0.314 \\
OmniGen2 & 0.558 & 0.387 & 0.34 & 0.533 & 0.7 & 0.352 & 0.46 \\
Show-o2 1.5B & 0.442 & 0.313 & 0.287 & 0.38 & 0.533 & 0.319 & 0.371 \\
Show-o2 & 0.617 & 0.427 & 0.373 & 0.473 & 0.692 & 0.395 & 0.479 \\
OmniGen2 & 0.558 & 0.387 & 0.34 & 0.533 & 0.7 & 0.352 & 0.46 \\
Emu3.5 & 0.35 & 0.24 & 0.34 & 0.32 & 0.242 & 0.271 & 0.292 \\
MMaDA & 0.292 & 0.347 & 0.213 & 0.313 & 0.333 & 0.257 & 0.289 \\
Show-o & 0.275 & 0.253 & 0.2 & 0.26 & 0.375 & 0.238 & 0.261 \\
Ovis-U1 & 0.567 & 0.393 & 0.333 & 0.4 & 0.658 & 0.367 & 0.437
\end{tabular}
}
\label{tab:sub-mmmu}
\end{table}

%% file: tab/app-sub-gedit-en-o.tex
\begin{table}[t]
\centering
\caption{GEdit-EN-Overall sub-score. 
}
\resizebox{\linewidth}{!}{
\begin{tabular}{l|cccccccccccc}
Model & \multicolumn{1}{l}{bg\_change} & \multicolumn{1}{l}{color} & \multicolumn{1}{l}{material} & \multicolumn{1}{l}{motion} & \multicolumn{1}{l}{ps\_human} & \multicolumn{1}{l}{style} & \multicolumn{1}{l}{subj-add} & \multicolumn{1}{l}{subj-rm} & \multicolumn{1}{l}{subj-repl} & \multicolumn{1}{l}{text} & \multicolumn{1}{l}{tone} & \multicolumn{1}{l}{Avg} \\ \midrule
Bagel SC & 7.45 & 7.075 & 6.975 & 5.4 & 5.271 & 6.233 & 7.283 & 7.175 & 7.167 & 6.838 & 6.6 & 6.679 \\
Bagel PQ & 7.35 & 7.075 & 6.725 & 7.175 & 7.043 & 5.833 & 7.7 & 7.211 & 7.283 & 7.535 & 6.55 & 7.044 \\
Bagel O & 7.264 & 6.653 & 6.448 & 5.128 & 5.046 & 5.803 & 7.214 & 6.743 & 6.861 & 6.368 & 6.304 & 6.348 \\
DeepGen SC & 7.775 & 7.9 & 7.625 & 7.675 & 7.114 & 5.983 & 7.967 & 7.596 & 7.833 & 7.242 & 7.175 & 7.444 \\
DeepGen PQ & 7.825 & 7.325 & 7.025 & 7.875 & 7.657 & 6.867 & 7.8 & 7.667 & 7.817 & 7.626 & 7.4 & 7.535 \\
DeepGen O & 7.78 & 7.44 & 7.286 & 7.745 & 7.2 & 6.287 & 7.869 & 7.409 & 7.737 & 6.836 & 7.052 & 7.331 \\
OmniGen2 SC & 7.45 & 7.575 & 6.15 & 6.55 & 4.886 & 6.75 & 7.083 & 6.228 & 6.783 & 5.404 & 6.5 & 6.487 \\
OmniGen2 PQ & 7.4 & 6.95 & 6.975 & 7.4 & 7.114 & 6.633 & 7.6 & 7.368 & 7.383 & 7.596 & 6.6 & 7.184 \\
OmniGen2 O & 7.195 & 6.992 & 5.901 & 6.56 & 4.885 & 6.476 & 6.98 & 5.963 & 6.475 & 5.246 & 6.277 & 6.268 \\
Emu3.5 SC & 7.975 & 7.875 & 7.842 & 7.775 & 7.1 & 7.217 & 7.983 & 6.895 & 7.814 & 8.722 & 6.875 & 7.643 \\
Emu3.5 PQ & 7.7 & 7.4 & 6.868 & 7.7 & 7.4 & 7.2 & 7.683 & 7.649 & 7.492 & 7.722 & 7.45 & 7.479 \\
Emu3.5 O & 7.836 & 7.634 & 7.339 & 7.737 & 7.248 & 7.208 & 7.832 & 7.262 & 7.651 & 8.206 & 7.157 & 7.556 \\
Omnigen2-SFT SC & 7.525 & 7.275 & 6.125 & 7.1 & 5.014 & 6.683 & 7.05 & 6 & 6.867 & 5.646 & 6.525 & 6.528 \\
Omnigen2-SFT PQ & 7.35 & 6.9 & 6.9 & 7.5 & 7.086 & 6.767 & 7.517 & 7.368 & 7.433 & 7.535 & 6.675 & 7.185 \\
Omnigen2-SFT O & 7.285 & 6.718 & 5.902 & 6.99 & 5.027 & 6.545 & 6.862 & 5.785 & 6.611 & 5.375 & 6.331 & 6.312 \\
bagel\_sft SC & 7.375 & 7.55 & 6.35 & 6.475 & 5.457 & 6.35 & 7.133 & 6.737 & 7.25 & 7.444 & 6.6 & 6.793 \\
bagel\_sft PQ & 7.35 & 7.025 & 6.7 & 7.05 & 7.043 & 5.9 & 7.667 & 7.246 & 7.167 & 7.576 & 6.55 & 7.025 \\
bagel\_sft O & 7.219 & 7.058 & 6.01 & 6.304 & 5.371 & 5.918 & 7.002 & 6.215 & 6.846 & 7.077 & 6.335 & 6.487 \\
bagel\_recA SC & 7.775 & 7.425 & 7.3 & 6.9 & 5.514 & 6.133 & 7.783 & 6.719 & 7.433 & 7.697 & 7.125 & 7.073 \\
bagel\_recA PQ & 7.35 & 7.1 & 6.925 & 7.15 & 6.9 & 6.217 & 7.683 & 7.456 & 7.183 & 7.566 & 6.725 & 7.114 \\
bagel\_recA O & 7.529 & 7.056 & 6.966 & 6.742 & 5.296 & 5.968 & 7.698 & 6.388 & 7.064 & 7.275 & 6.769 & 6.795 \\
bagel\_IRG SC & 6.6 & 7.875 & 5.675 & 6.725 & 5.771 & 6.6 & 7.7 & 5.947 & 7.233 & 5.808 & 6.85 & 6.617 \\
bagel\_IRG PQ & 7.325 & 7.075 & 7.05 & 7.4 & 6.729 & 6.4 & 7.8 & 7.719 & 7.383 & 7.556 & 7.3 & 7.249 \\
bagel\_IRG O & 6.469 & 7.299 & 5.469 & 6.645 & 5.654 & 6.256 & 7.661 & 5.846 & 6.951 & 5.714 & 6.896 & 6.442 \\
bagel\_unigame SC & 7.475 & 7.575 & 5.900 & 6.425 & 5.886 & 6.433 & 7.167 & 7.105 & 7.617 & 7.242 & 6.475 & 6.845 \\
bagel\_unigame PQ & 7.300 & 6.925 & 6.725 & 7.025 & 6.829 & 5.833 & 7.667 & 7.281 & 7.083 & 7.545 & 6.550 & 6.978 \\
bagel\_unigame O  & 7.243 & 6.999 & 5.610 & 6.036 & 5.564 & 5.975 & 7.120 & 6.564 & 7.171 & 6.825 & 6.180 & 6.481 \\
omnigen2\_sft SC & 7.525 & 7.275 & 6.125 & 7.1 & 5.014 & 6.683 & 7.05 & 6 & 6.867 & 5.646 & 6.525 & 6.528 \\
omnigen2\_sft PQ & 7.35 & 6.9 & 6.9 & 7.5 & 7.086 & 6.767 & 7.517 & 7.368 & 7.433 & 7.535 & 6.675 & 7.185 \\
omnigen2\_sft O & 7.285 & 6.718 & 5.902 & 6.99 & 5.027 & 6.545 & 6.862 & 5.785 & 6.611 & 5.375 & 6.331 & 6.312 \\
ovis\_u1 SC & 7.775 & 7.85 & 7.175 & 7.375 & 7.029 & 6.883 & 7.85 & 7.14 & 8.033 & 5.535 & 7.475 & 7.284 \\
ovis\_u1 PQ & 7.75 & 7.3 & 7.25 & 7.25 & 7.543 & 6.883 & 7.75 & 7.807 & 7.667 & 7.424 & 7.475 & 7.464 \\
ovis\_u1 O & 7.746 & 7.42 & 7.024 & 7.181 & 7.07 & 6.748 & 7.773 & 6.981 & 7.829 & 5.367 & 7.336 & 7.134
\end{tabular}
}
\label{tab:sub-gedit-en-o}
\end{table}

%% file: tab/app-sub-gedit-en-in.tex
\begin{table}[t]
\centering
\caption{GEdit-EN-Intersection sub-score. 
}
\resizebox{\textwidth}{!}{
\begin{tabular}{l|cccccccccccc}
Model & \multicolumn{1}{l}{bg\_change} & \multicolumn{1}{l}{color} & \multicolumn{1}{l}{material} & \multicolumn{1}{l}{motion} & \multicolumn{1}{l}{ps\_human} & \multicolumn{1}{l}{style} & \multicolumn{1}{l}{subj-add} & \multicolumn{1}{l}{subj-rm} & \multicolumn{1}{l}{subj-repl} & \multicolumn{1}{l}{text} & \multicolumn{1}{l}{tone} & \multicolumn{1}{l}{Avg} \\ \midrule
Bagel SC & 7.241 & 7.353 & 7.071 & 4.909 & 5.61 & 6.146 & 7.342 & 7.19 & 7.174 & 6.864 & 7.08 & 6.726 \\
Bagel PQ & 7.172 & 6.882 & 6.714 & 7.227 & 7.195 & 5.833 & 7.658 & 7.238 & 7.217 & 7.519 & 6.64 & 7.027 \\
Bagel O & 7.021 & 6.872 & 6.493 & 4.662 & 5.401 & 5.741 & 7.264 & 6.784 & 6.813 & 6.391 & 6.789 & 6.384 \\
DeepGen SC & 7.862 & 8.059 & 7.607 & 7.682 & 7.537 & 5.771 & 7.974 & 7.881 & 7.891 & 7.136 & 7.32 & 7.52 \\
DeepGen PQ & 7.828 & 7.176 & 7.143 & 8 & 7.805 & 6.812 & 7.711 & 7.857 & 7.87 & 7.642 & 7.68 & 7.593 \\
DeepGen O & 7.839 & 7.548 & 7.351 & 7.807 & 7.531 & 6.134 & 7.823 & 7.753 & 7.782 & 6.768 & 7.323 & 7.423 \\
OmniGen2 SC & 7.276 & 7.941 & 6.071 & 6.045 & 5.122 & 6.562 & 6.763 & 6.31 & 6.783 & 5.531 & 6.72 & 6.466 \\
OmniGen2 PQ & 7.276 & 6.765 & 7.107 & 7.773 & 7.415 & 6.75 & 7.5 & 7.476 & 7.522 & 7.593 & 6.68 & 7.26 \\
OmniGen2 O & 6.96 & 7.271 & 5.843 & 6.187 & 5.245 & 6.453 & 6.656 & 6.049 & 6.541 & 5.387 & 6.5 & 6.281 \\
Emu3.5 SC & 7.975 & 7.875 & 7.842 & 7.775 & 7.1 & 7.217 & 7.983 & 6.895 & 7.814 & 8.722 & 6.875 & 7.643 \\
Emu3.5 PQ & 7.7 & 7.4 & 6.868 & 7.7 & 7.4 & 7.2 & 7.683 & 7.649 & 7.492 & 7.722 & 7.45 & 7.479 \\
Emu3.5 O& 7.836 & 7.634 & 7.339 & 7.737 & 7.248 & 7.208 & 7.832 & 7.262 & 7.651 & 8.206 & 7.157 & 7.556 \\
Omnigen2-SFT SC & 7.379 & 7.441 & 6.036 & 6.818 & 5.439 & 6.5 & 6.737 & 6.262 & 6.87 & 5.889 & 6.76 & 6.557 \\
Omnigen2-SFT PQ & 7.207 & 6.706 & 7.107 & 7.864 & 7.366 & 6.729 & 7.421 & 7.476 & 7.5 & 7.506 & 6.88 & 7.251 \\
Omnigen2-SFT O & 7.084 & 6.762 & 5.866 & 6.95 & 5.551 & 6.407 & 6.51 & 6.025 & 6.652 & 5.616 & 6.631 & 6.369 \\
bagel\_sft SC & 7.138 & 7.441 & 6.893 & 6.227 & 5.878 & 6.292 & 7.105 & 7.095 & 7.348 & 7.84 & 7.08 & 6.94 \\
bagel\_sft PQ & 7.172 & 6.853 & 6.821 & 7.227 & 6.976 & 5.896 & 7.658 & 7.238 & 7.152 & 7.556 & 6.6 & 7.014 \\
bagel\_sft O & 6.959 & 6.878 & 6.547 & 6.269 & 5.747 & 5.875 & 6.958 & 6.472 & 6.956 & 7.435 & 6.747 & 6.622 \\
bagel\_recA SC & 7.69 & 7.382 & 7.643 & 6.409 & 5.683 & 6.083 & 7.711 & 7.214 & 7.37 & 7.691 & 7.6 & 7.134 \\
bagel\_recA PQ & 7.276 & 6.941 & 7.036 & 7.591 & 6.976 & 6.271 & 7.632 & 7.571 & 7.087 & 7.568 & 7 & 7.177 \\
bagel\_recA O & 7.44 & 6.924 & 7.304 & 6.56 & 5.567 & 5.946 & 7.619 & 6.885 & 6.922 & 7.303 & 7.267 & 6.885 \\
bagel\_IRG SC & 6.724 & 8.059 & 5.607 & 6.273 & 5.976 & 6.625 & 7.895 & 6.548 & 7.391 & 5.864 & 7.08 & 6.731 \\
bagel\_IRG PQ & 7.172 & 6.912 & 6.75 & 7.409 & 6.488 & 6.375 & 7.737 & 7.881 & 7.522 & 7.531 & 7.24 & 7.183 \\
bagel\_IRG O & 6.551 & 7.396 & 5.331 & 6.23 & 5.775 & 6.222 & 7.775 & 6.436 & 7.184 & 5.713 & 7.061 & 6.516 \\
bagel\_unigame SC & 7.241 & 7.441 & 6.679 & 5.909 & 6.341 & 6.396 & 6.711 & 7.476 & 7.609 & 7.074 & 6.600 & 6.862 \\
bagel\_unigame PQ & 7.138 & 6.735 & 6.821 & 6.864 & 7.000 & 5.854 & 7.579 & 7.357 & 7.022 & 7.531 & 6.720 & 6.966 \\
bagel\_unigame O  & 6.993 & 6.794 & 6.306 & 5.487 & 6.111 & 5.960 & 6.678 & 6.897 & 7.106 & 6.669 & 6.293 & 6.481 \\
omnigen2\_sft SC & 7.379 & 7.441 & 6.036 & 6.818 & 5.439 & 6.5 & 6.737 & 6.262 & 6.87 & 5.889 & 6.76 & 6.557 \\
omnigen2\_sft PQ & 7.207 & 6.706 & 7.107 & 7.864 & 7.366 & 6.729 & 7.421 & 7.476 & 7.5 & 7.506 & 6.88 & 7.251 \\
omnigen2\_sft O & 7.084 & 6.762 & 5.866 & 6.95 & 5.551 & 6.407 & 6.51 & 6.025 & 6.652 & 5.616 & 6.631 & 6.369 \\
ovis\_u1 SC & 7.862 & 8.029 & 7.357 & 8 & 7.366 & 6.729 & 7.842 & 6.976 & 8.087 & 5.383 & 7.68 & 7.392 \\
ovis\_u1 PQ & 7.724 & 7.176 & 7.107 & 7.727 & 7.585 & 6.854 & 7.684 & 7.929 & 7.739 & 7.432 & 7.52 & 7.498 \\
ovis\_u1 O & 7.785 & 7.539 & 7.113 & 7.844 & 7.33 & 6.638 & 7.723 & 6.876 & 7.905 & 5.232 & 7.55 & 7.231
\end{tabular}
}
\label{tab:sub-gedit-en-in}
\end{table}

%% file: tab/app-sub-realunify.tex
\begin{table}[t]
\centering
\caption{RealUnify UEG-score. 
}
\renewcommand{\arraystretch}{1.2}
\resizebox{.7\linewidth}{!}{
\begin{tabular}{l|ccccccc}
\hline
\multirow{2}{*}{Model} & \multicolumn{7}{c}{RealUnify(UEG)} \\ \cline{2-8} 
 & WK & CR & LR & MR & SR & CTI & Total \\ \midrule
Bagel(14B) & 0.38 & 0.31 & 0.12 & 0.21 & 0.19 & 0.08 & 0.22 \\
Janus-Pro(7B) & 0.26 & 0.36 & 0.07 & 0.08 & 0.14 & 0.02 & 0.16 \\
Janus(1.5B) & 0.09 & 0.05 & 0.00 & 0.08 & 0.11 & 0.00 & 0.06 \\
Janus-flow(1.5B) & 0.14 & 0.17 & 0.05 & 0.15 & 0.17 & 0.02 & 0.12 \\
Omnigen2(7B) & 0.36 & 0.25 & 0.12 & 0.23 & 0.21 & 0.18 & 0.23 \\
Show-o2(7B) & 0.22 & 0.25 & 0.13 & 0.22 & 0.20 & 0.10 & 0.19 \\
Show-o2(1.5B) & 0.15 & 0.23 & 0.12 & 0.19 & 0.17 & 0.07 & 0.16 \\
Show-o(1.3B) & 0.13 & 0.18 & 0.09 & 0.19 & 0.16 & 0.01 & 0.13 \\
Ovis-u1(3.6B) & 0.28 & 0.36 & 0.17 & 0.25 & 0.17 & 0.05 & 0.21 \\
MMaDA(8B) & 0.19 & 0.14 & 0.05 & 0.19 & 0.15 & 0.00 & 0.12 \\ \hline
\end{tabular}
}
\label{tab:sub-realunify}
\end{table}

%% file: tab/app-uni-mmmu.tex
\begin{table}[t]
\centering
\caption{Uni-mmmu sub-score. 
} 
\resizebox{\textwidth}{!}{
\begin{tabular}{l|cccccccccccccc}
\hline
Model & \multicolumn{2}{c}{und+edit} & \multicolumn{2}{c}{und+edit} & \multicolumn{2}{c}{und+edit} & \multicolumn{2}{c}{und+edit} & \multicolumn{3}{c}{und+gen} & \multicolumn{3}{c}{und+gen} \\ \hline
 & Jig. I & Jig. T & Maze I & Maze T & Slid. I & Slid. T & Geo I & Geo T & Sci. R. & Sci. T & Sci. I & C. T & C. S & C. P \\ \hline
bagel(w/o think) & 0.66 & 0.553 & 0.004 & 0.101 & 0 & 0.05 & 0.05 & 0.143 & 0.592 & 0.522 & 0.185 & 0.115 & 0.375 & 0.275 \\ 
Janus\_pro & - & - & - & - & - & - & - & - & 29.3 & 25.5 & 0 & 1.5 & 3.7 & 3.4 \\ \hline
\end{tabular}
}
\label{tab:uni-mmmu}
\end{table}

%% file: checklist.tex
\section*{NeurIPS Paper Checklist}

\begin{enumerate}

\item {\bf Claims}
    \item[] Question: Do the main claims made in the abstract and introduction accurately reflect the paper's contributions and scope?
    \item[] Answer: \answerYes{} 
    \item[] Justification: Yes, the main claims made in the abstract and introduction accurately reflect the paper's contributions and scope.

\item {\bf Limitations}
    \item[] Question: Does the paper discuss the limitations of the work performed by the authors?
    \item[] Answer: \answerYes{} 
    \item[] Justification: We discuss the limitations in \Cref{sec-conclusion}.

\item {\bf Theory assumptions and proofs}
    \item[] Question: For each theoretical result, does the paper provide the full set of assumptions and a complete (and correct) proof?
    \item[] Answer: \answerNA{} 
    \item[] Justification: The paper does not present theoretical theorems or proof claims.

    \item {\bf Experimental result reproducibility}
    \item[] Question: Does the paper fully disclose all the information needed to reproduce the main experimental results of the paper to the extent that it affects the main claims and/or conclusions of the paper (regardless of whether the code and data are provided or not)?
    \item[] Answer: \answerYes{} 
    \item[] Justification: We include all the details in the code released and readme file.

\item {\bf Open access to data and code}
    \item[] Question: Does the paper provide open access to the data and code, with sufficient instructions to faithfully reproduce the main experimental results, as described in supplemental material?
    \item[] Answer: \answerYes{} 
    \item[] Justification: We released our codebase.

\item {\bf Experimental setting/details}
    \item[] Question: Does the paper specify all the training and test details (e.g., data splits, hyperparameters, how they were chosen, type of optimizer) necessary to understand the results?
    \item[] Answer: \answerYes{} 
    \item[] Justification: We specify it in paper and the code released.

\item {\bf Experiment statistical significance}
    \item[] Question: Does the paper report error bars suitably and correctly defined or other appropriate information about the statistical significance of the experiments?
    \item[] Answer: \answerNo{} 
    \item[] Justification: We report results based on single-run evaluations using fixed checkpoints and standard inference settings. Due to the high computational cost of running large-scale multimodal models and post-training variants, we do not include multiple random seeds or confidence intervals.

\item {\bf Experiments compute resources}
    \item[] Question: For each experiment, does the paper provide sufficient information on the computer resources (type of compute workers, memory, time of execution) needed to reproduce the experiments?
    \item[] Answer: \answerYes{} 
    \item[] Justification: We reported it in the paper and appendix.
    
\item {\bf Code of ethics}
    \item[] Question: Does the research conducted in the paper conform, in every respect, with the NeurIPS Code of Ethics \url{https://neurips.cc/public/EthicsGuidelines}?
    \item[] Answer: \answerYes{} 
    \item[] Justification: We confirm that the work follows the NeurIPS Code of Ethics.

\item {\bf Broader impacts}
    \item[] Question: Does the paper discuss both potential positive societal impacts and negative societal impacts of the work performed?
    \item[] Answer: \answerYes{} 
    \item[] Justification: See details in \Cref{sec-appendix-impact}.
    
\item {\bf Safeguards}
    \item[] Question: Does the paper describe safeguards that have been put in place for responsible release of data or models that have a high risk for misuse (e.g., pre-trained language models, image generators, or scraped datasets)?
    \item[] Answer: \answerNA{} 
    \item[] Justification: The submission does not release a new high-risk pretrained model, image generator, or scraped image dataset.

\item {\bf Licenses for existing assets}
    \item[] Question: Are the creators or original owners of assets (e.g., code, data, models), used in the paper, properly credited and are the license and terms of use explicitly mentioned and properly respected?
    \item[] Answer: \answerYes{} 
    \item[] Justification: Existing datasets, benchmarks, and models are credited.

\item {\bf New assets}
    \item[] Question: Are new assets introduced in the paper well documented and is the documentation provided alongside the assets?
    \item[] Answer: \answerYes{} 
    \item[] Justification: New assets introduced in the paper well documented.

\item {\bf Crowdsourcing and research with human subjects}
    \item[] Question: For crowdsourcing experiments and research with human subjects, does the paper include the full text of instructions given to participants and screenshots, if applicable, as well as details about compensation (if any)? 
    \item[] Answer: \answerNA{} 
    \item[] Justification: The work does not involve crowdsourcing or human-subject experiments.

\item {\bf Institutional review board (IRB) approvals or equivalent for research with human subjects}
    \item[] Question: Does the paper describe potential risks incurred by study participants, whether such risks were disclosed to the subjects, and whether Institutional Review Board (IRB) approvals (or an equivalent approval/review based on the requirements of your country or institution) were obtained?
    \item[] Answer: \answerNA{} 
    \item[] Justification: The work does not involve crowdsourcing or human-subject experiments, so IRB approval is not applicable.

\item {\bf Declaration of LLM usage}
    \item[] Question: Does the paper describe the usage of LLMs if it is an important, original, or non-standard component of the core methods in this research? Note that if the LLM is used only for writing, editing, or formatting purposes and does \emph{not} impact the core methodology, scientific rigor, or originality of the research, declaration is not required.
    \item[] Answer: \answerYes{} 
    \item[] Justification: Yes, in Appendix \Cref{sec-app-llm}

\end{enumerate}